\documentclass[journal=jacsat,manuscript=article]{achemso}

\usepackage[version=3]{mhchem} 
\usepackage{mciteplus}
\mciteErrorOnUnknownfalse
\usepackage{natbib}
\usepackage{caption}
\usepackage{subcaption}

\DeclareCaptionFont{eightpt}{\fontsize{8pt}{10pt}\selectfont}

\usepackage{placeins}

\usepackage{tabularx}

\usepackage{xurl}   
\usepackage{hyperref}

\usepackage[usenames,dvipsnames]{xcolor}
\usepackage{float} 
\usepackage{gensymb}
\usepackage{amsmath}
\usepackage{multirow}

\usepackage{amssymb}  
\usepackage{amsfonts}  

\usepackage{graphicx}
\usepackage{xcolor}

\usepackage{newunicodechar}
\newunicodechar{✓}{\checkmark}

\author{Charles Rhys Campbell}
\email{crhysc@gmail.com}
\affiliation
{Department of Physics and Astronomy, West Virginia University, Morgantown, WV 26506, USA}

\author{Aldo H. Romero}
\email{alromero@mail.wvu.edu}
\affiliation
{Department of Physics and Astronomy, West Virginia University, Morgantown, WV 26506, USA}
\author{Kamal Choudhary}
\email{kchoudh2@jhu.edu}
\affiliation
{Department of Materials Science and Engineering, Whiting School of Engineering, The Johns Hopkins University, Baltimore, MD 21218, USA}
\altaffiliation
{Department of Electrical and Computer Engineering, Whiting School of Engineering, The Johns Hopkins University, Baltimore, MD 21218, USA}

\title{AtomBench: A Benchmarking Framework for Generative Crystal Reconstruction Models in Conventional Superconductors}

\abbreviations{IR,NMR,UV}

\begin{document}


\begin{abstract}
 {
 A key question in benchmarking generative crystal reconstruction models is how the amount and type of crystallographic information provided to a generative model affects its ability to reconstruct atomic structures. Yet such comparisons often overlook the fact that models receive unequal information about the target during reconstruction, thereby confounding architectural conclusions. We present AtomBench, an extensible, model-agnostic framework for comparing generative models on a well-defined crystal reconstruction task (rather than \textit{de novo} generation), which we here apply to conventional superconductors. We train and evaluate four models, AtomGPT, CDVAE, FlowMM, and MatterGen, on the JARVIS Supercon-3D and Alexandria DS-A/B datasets, grouping them by the information each accesses at inference. Reconstruction fidelity is measured by the Kullback-Leibler divergence (KLD) and mean absolute error (MAE) of lattice parameters and the root-mean-squared displacement (RMSD) of atomic coordinates. {We further introduce the continuous corrected RMSD (ccRMSD), a continuous measure of local geometric fidelity defined for every structure in the test set.} MatterGen achieves the best atomic-coordinate reconstruction, followed by AtomGPT, while CDVAE reconstructs lattices most accurately, and FlowMM is the least accurate but fastest overall. We find that conditioning on critical temperature T$_c$ does not consistently improve fidelity. {We also release AtomBench as an open-source Python package that reproduces all reported reconstruction metrics, figures, and tables from one or more benchmark files and supports direct submission to the JARVIS-Leaderboard. Any inverse model emitting crystal reconstructions can be benchmarked with \texttt{atombench}, and we encourage community use.}
} \url{https://github.com/atomgptlab/atombench}.
\end{abstract}

\section{Introduction}

Electrons in crystalline solids can organize into remarkable collective states, with high-temperature superconductivity as a prime example at the forefront of condensed-matter physics and materials science. The conventional \textit{in-silico} discovery pipeline relies heavily on density functional theory (DFT), which computes structure-property relationships from first principles and remains the most trusted approach for conventional, electron-phonon-mediated superconductors, where the superconductivity mechanism is fairly understood~\cite{giustino2017electron, oliveira1988density, luders2005ab, gross2013density, choudhary2021atomistic}. DFT is, however, computationally expensive, with the wall-clock time scaling roughly as $\mathcal{O}(N^3)$ with system size, which significantly limits its throughput. Machine learning offers a complementary route: trained on DFT-computed corpora, generative models inherit DFT's physical and chemical constraints while sharply lowering the cost of proposing candidate structures~\cite{choudhary2022recent}. Such models fall into two broad families. The first is forward design, which focuses on predicting properties from a fully specified structure. The second is inverse design, which maps a target property back to candidate structures. The latter is the substantially harder, ill-posed problem this work addresses. An extended treatment of DFT-based superconductor discovery and the forward/inverse-design formalism is provided in the Appendix.

Although many forward and inverse models exist, this study focuses exclusively on benchmarking inverse models.
A variety of architectural families have been proposed and progressively refined in the literature, yielding increasingly sophisticated and accurate models. Among these, diffusion models have emerged as a widely used paradigm, in which a learned denoising process transforms random noise into a physically plausible crystal structure. Example diffusion models include DiffCSP\cite{Jiao2023DiffCSP}, SymmCD\cite{Levy2025SymmCD}, CDVAE\cite{Xie2022CDVAE}, and MatterGen\cite{Zeni2025MatterGen}, the latter two of which we benchmark in this study. Another increasingly prominent paradigm is the use of transformers to generate stable inorganic materials as text. These models can be trained on many types of data, from raw structure/property pairs to large corpora of materials science literature, and examples of such workflows are LLaMat~\cite{Mishra2024LLaMat}, CrystaLLM~\cite{Antunes2024CrystaLLM}, MatExpert~\cite{Ding2024MatExpert}, and AtomGPT~\cite{Choudhary2023AtomGPT}, the latter of which we choose to benchmark in this study. Finally, conditional flow-matching is another increasingly represented architecture. Here, a transport ODE defining a learned flow evolves random noise into plausible crystal structures over a high-dimensional manifold of crystals. Examples include CrystalFlow~\cite{Luo2024CrystalFlow} and FlowMM~\cite{Miller2024FlowMM}, the latter of which we benchmark in this study. More broadly, the generative-transformer paradigm has been extended beyond crystal generation to related materials characterization and design tasks, including atomic-structure determination from X-ray diffraction patterns~\cite{Choudhary2025DiffractGPT} and electron-microscopy images~\cite{Choudhary2025MicroscopyGPT}, as well as conversational and agentic assistants that orchestrate materials workflows~\cite{Choudhary2025CME, Lee2025AGAPI}.

In their associated publications, models are almost always benchmarked and tested using a variety of statistical measures. Generally, two tasks are evaluated when benchmarking inverse models. The first benchmarking task is \textit{de novo} generation, in which the stability, uniqueness, and novelty (S.U.N.) of the generated structures are assessed. Most frequently, the S.U.N. rate is reported as a single scalar value for a given experiment, and it represents the number of structures that satisfy a chosen S.U.N. threshold divided by the total number of structures generated by the model. For the AtomBench benchmarking framework, we focus only on the second task: crystal structure prediction (CSP), also known as crystal reconstruction. The goal of CSP is to evaluate the fidelity with which a trained inverse model is able to reproduce a distribution of held-out crystals from the set of training crystals. Most commonly, papers focus on the percentage of predicted crystals that match (within a chosen tolerance) the crystals in the held-out set, as well as the root-mean-squared error between predicted and ground-truth atomic coordinates. The divergence between predicted lattice parameters and ground-truth lattice parameters is also benchmarked, but it is relatively rare among methods papers\cite{jiao2023crystal}.

{A subtle but consequential limitation of this standard protocol is that both the match rate and the matched RMSD depend on a hard structure-matching tolerance, typically evaluated with the \texttt{pymatgen} \texttt{StructureMatcher}. Under this scheme, a predicted crystal either matches its ground-truth counterpart within fixed length, angle, and site tolerances, or is discarded; the reported coordinate error is averaged only over the surviving matches. The match/no-match decision is discontinuous in the atomic coordinates, so an arbitrarily small perturbation of a near-threshold structure can switch its contribution on or off, and the resulting statistic is conditioned on precisely the structures that were easiest to match. For inverse-model benchmarking, this is especially problematic: when a model matches only a small fraction of the test set, the matched root mean squared displacement (RMSD) is computed over a small, favorably selected subsample and can appear deceptively good while the bulk of the reconstructions are ignored. This limitation has motivated a broader effort in the crystallography literature to develop continuous crystal distance functions. Prominent examples include the average minimum distance (AMD) and the related pointwise distance distribution (PDD), provably continuous (Lipschitz) invariants of periodic point sets that have been used to continuously map structural databases as large as the Cambridge Structural Database~\cite{widdowson2022amd, widdowson2022pdd, widdowson2024csdmaps}. We adopt this continuous-distance viewpoint to define a complementary, match-rate-independent coordinate metric (ccRMSD) in the Methods.}

On a different note, AtomBench addresses a potential confounding factor in the current literature: inverse models are often assumed to have access to the same information about the target crystal prior to making a prediction. The goal of benchmarking crystal structure prediction is to measure how well training enables the model to encode the statistical structure of the training set, and a given model's performance can be artificially inflated if it has more information about the target crystal than the model it is being compared with. Moreover, this challenges the notion that a model generalizes better if its improved performance results from greater information about the target crystal. Later in this paper, we formalize this concept as the conditional entropy of the model's inference. As computing literal conditional entropies is generally intractable, we propose using qualitative conditional entropy inequalities to distinguish models that provide different amounts of information about the target crystals during reconstruction.

This information-discrepancy view also motivates a direct test of scalar property conditioning: two of the four models we benchmark (AtomGPT and MatterGen) natively condition on the superconducting critical temperature T$_c$, so we evaluate each with and without it. We find that T$_c$ conditioning does not consistently improve reconstruction fidelity, suggesting that a single scalar property contributes little structural information beyond composition (see Results and Discussion).


 \section{Methods}
\begin{figure}[h]
\centering
  \includegraphics[width=0.65\linewidth]{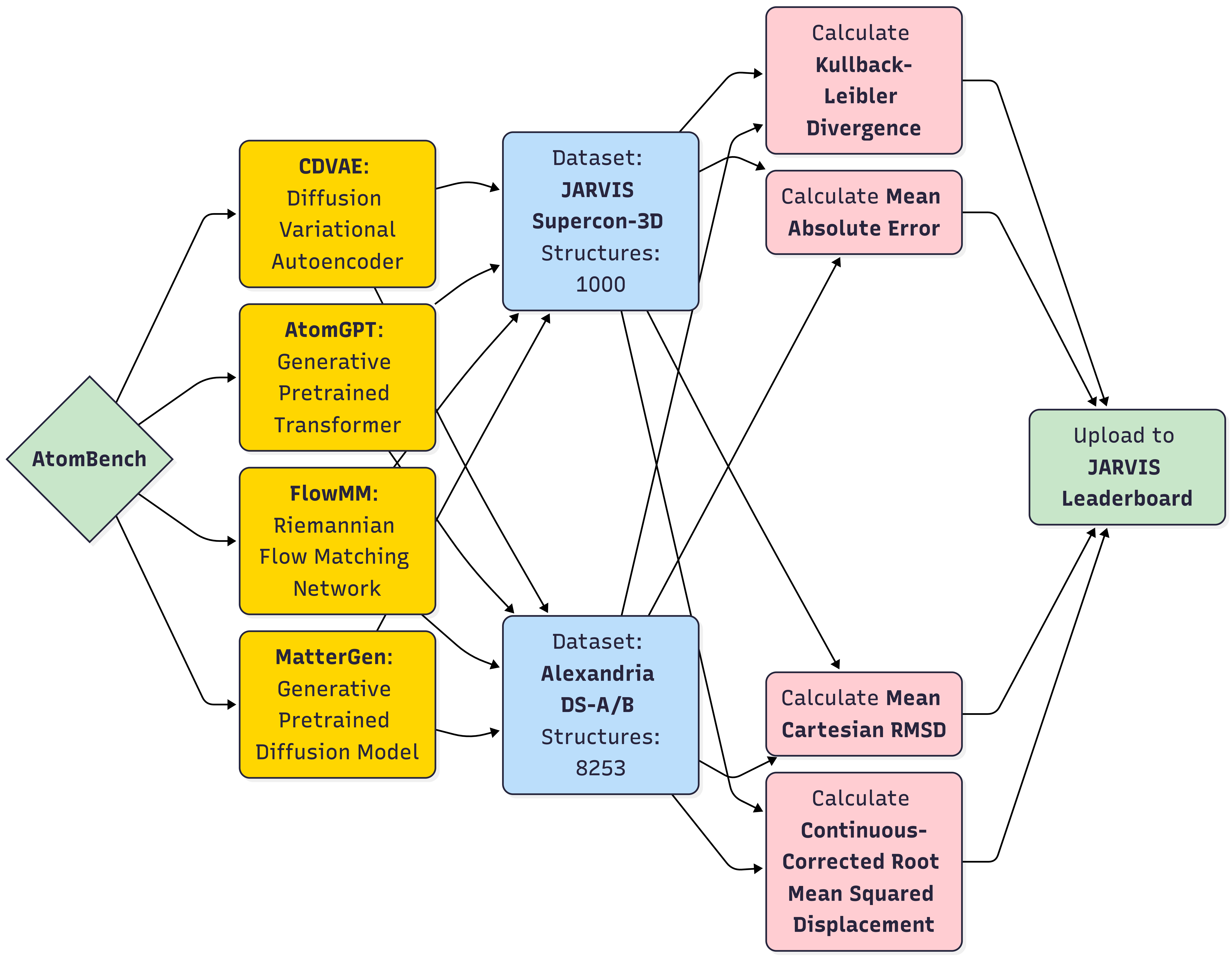}
  \caption{
    Diagram showing the inverse model benchmarking study design. We compare {four} generative AI inverse models-AtomGPT~\cite{Choudhary2023AtomGPT}, a language model; CDVAE\cite{Xie2022CDVAE}, a diffusion variational autoencoder; FlowMM\cite{Miller2024FlowMM}, a flow-matching network{; and MatterGen\cite{Zeni2025MatterGen}, a joint-diffusion model}-each separately trained on two superconductivity DFT datasets (JARVIS Supercon-3D\cite{Choudhary_2022} and Alexandria DS-A/B\cite{CerqueiraHydrideSupercons}){, with T$_c$-conditioned and unconditioned variants for the two models that support property conditioning,} for a total of {twelve} benchmarks. Ten percent of each dataset is held out to assess reconstruction performance, which we quantify statistically before submitting these results to the JARVIS-Leaderboard. {The models shown are illustrative; AtomBench is model-agnostic, and any inverse model that produces crystal reconstructions can be substituted into this workflow.}
  }

  \label{fgr:workflow_diagram}
\end{figure}

\begin{figure}[ht]
  \centering
  \begin{subfigure}[t]{0.5\textwidth}
    \centering
    \includegraphics[width=\linewidth]{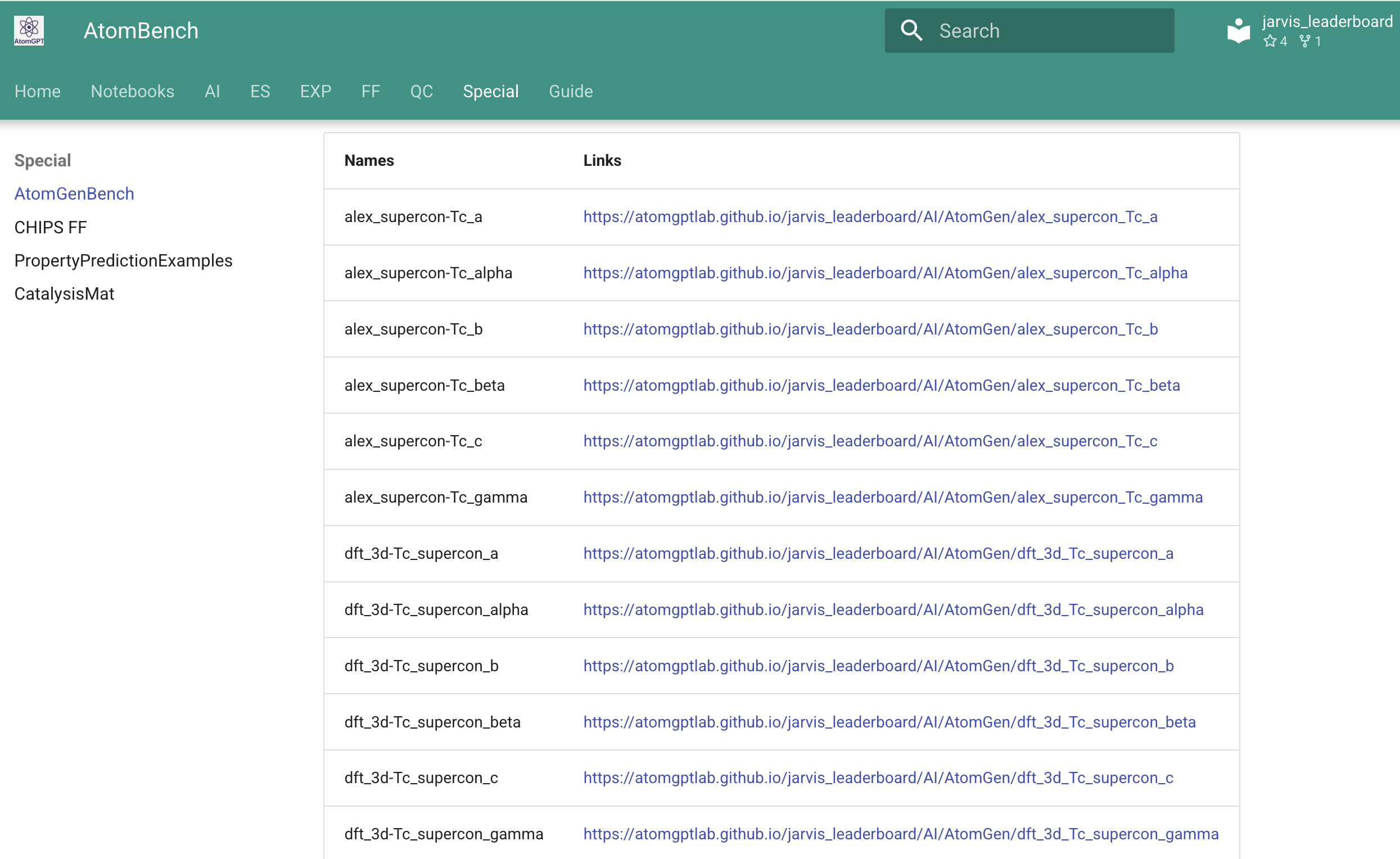}
    \subcaption{}
    \label{fig:jarvis_leaderboard_screenshot}
  \end{subfigure}
  \hspace{1em}
  \begin{subfigure}[t]{0.5\textwidth}
    \centering
    \includegraphics[width=\linewidth]{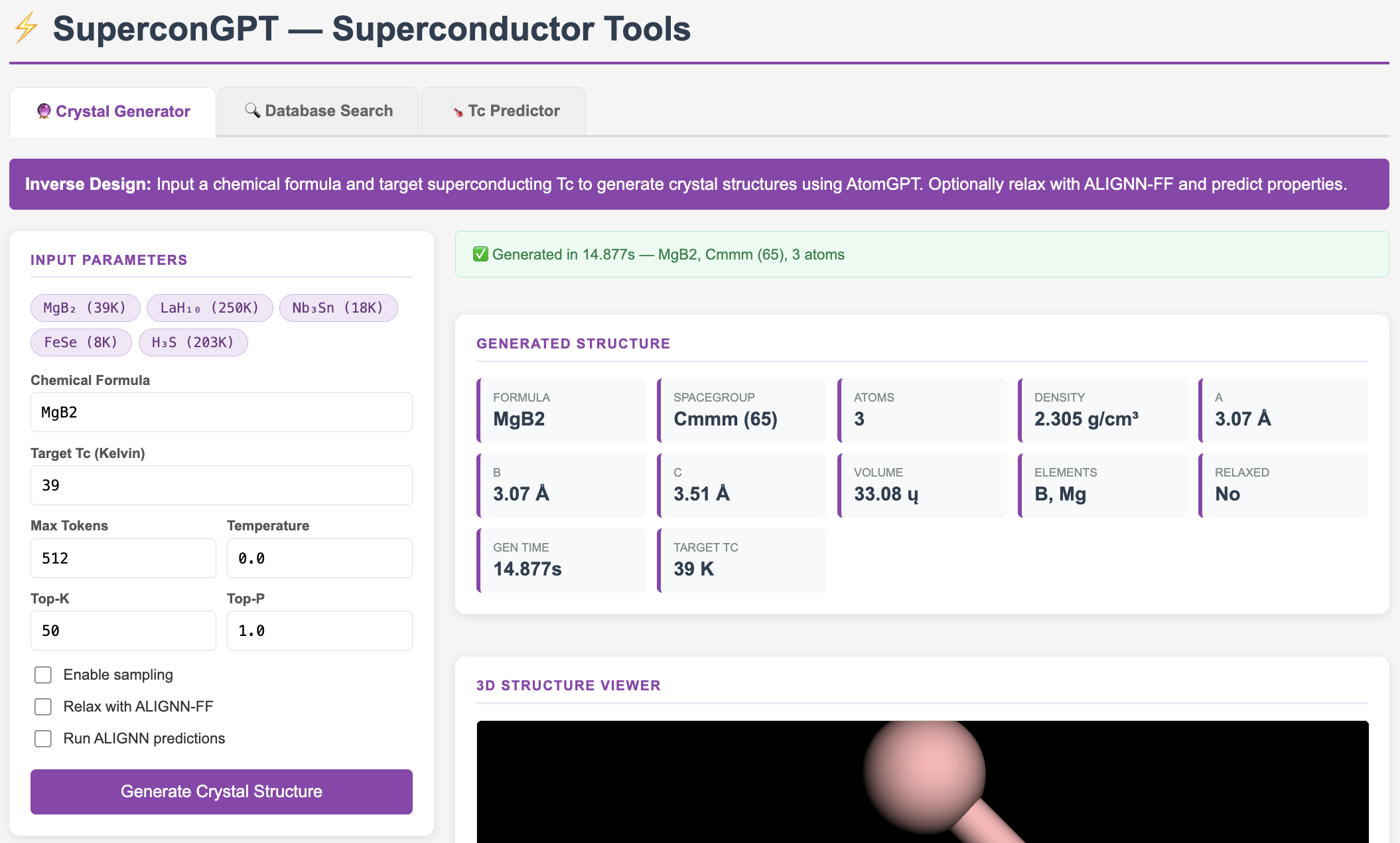}
    \subcaption{}
    \label{fig:atomgpt_org_supercon_screenshot}
  \end{subfigure}

  \caption{
  Screenshot of JARVIS Leaderboard AtomBench entries and Screenshot of AtomGPT.org. (a) Shows the AtomBench entry on the JARVIS Leaderboard~ \url{https://atomgptlab.github.io/jarvis_leaderboard/Special/AtomGenBench/}. Each hyperlink corresponds to a webpage that contains reconstruction performance for all models on a given dataset and lattice parameter. (b) Shows SuperconGPT, a web-app hosted on \url{https://atomgpt.org/supercon} that lets users generate T$_c$-conditioned crystal structures in real time. Currently, AtomGPT is the only model implemented in this web app, and the capabilities of CDVAE, FlowMM, and MatterGen are under development.
  }
  \label{fig:screenshots}
\end{figure}

A schematic of the current work design is summarized in Figure~\ref{fgr:workflow_diagram}. In this work, we performed a systematic comparison of {four} inverse design models: AtomGPT\cite{Choudhary2023AtomGPT}, a large language model; Crystal Diffusion Variational Autoencoder (CDVAE)\cite{Xie2022CDVAE}, a diffusion variational autoencoder model; FlowMM\cite{Miller2024FlowMM}, a Riemannian flow-matching network{; and MatterGen\cite{Zeni2025MatterGen}, a joint-diffusion model}. Two datasets were used in this study for model training and testing (labeled JARVIS Supercon-3D\cite{Choudhary_2022} and Alexandria DS-A/B\cite{CerqueiraHydrideSupercons}), and they are both comprised of DFT calculations. Each dataset comprises input-output pairs, where the input is a parameterized graph representation of the material and the output is its superconducting transition temperature computed from DFT. A separate instance of each model was trained on each dataset{, and for the two models that support T$_c$ conditioning (AtomGPT and MatterGen) we additionally trained T$_c$-conditioned variants,} resulting in {twelve} trained models in total. For each of the {twelve} model instances, 10\% of the training data was withheld from training, and each model was tasked with reproducing the unseen 10\% of its corresponding dataset. We then performed a statistical assessment of each model’s ability to reconstruct the held-out data, and the resulting performance metrics were uploaded to the JARVIS-Leaderboard~\cite{JARVIS-Leaderboard}, an established and community-driven benchmarking platform for materials AI models, available at: \url{https://atomgptlab.github.io/jarvis_leaderboard/Special/AtomGenBench/}. A web-app for generative superconductor design is also available at: \url{https://atomgpt.org/supercon}. Figure~\ref{fig:screenshots} shows these platforms visually.

{The analysis pipeline used to produce the reconstruction-accuracy results in this study is released as an open-source, pip-installable Python package, \texttt{atombench}. To benchmark a model, a user first generates a reconstruction analysis file formatted as a CSV containing \texttt{crystal id}, \texttt{target crystal}, and \texttt{predicted crystal} columns, where \texttt{target crystal} and \texttt{predicted crystal} columns hold the ground-truth and reconstructed structures as CIF or POSCAR-formatted strings. Then, the user runs a single command, and every benchmark the user supplies as input is overlaid in shared figures and pooled into a single comparison table. \texttt{atombench} computes the reconstruction-accuracy metrics defined in this section (lattice-parameter KLD and MAE, atomic-coordinate RMSD and ccRMSD, match rate, and per-crystal-system MAE), renders the corresponding figures presented in the Results, and writes the summary metrics table in three formats (CSV, JSON, and LaTeX) so that the LaTeX table can be pasted directly into a manuscript. The same functionality is exposed as an importable Python module for use in Python scripts. Finally, \texttt{atombench} lets users push their reconstruction results to the JARVIS-Leaderboard directly from Python or the command line, mirroring the submission workflow used in this study. We emphasize that the four models benchmarked here are a demonstration of the framework rather than a fixed set: \texttt{atombench} is model-agnostic, and any inverse model that produces crystal reconstructions can be substituted into this workflow and compared on equal footing.}

Next, the scope and limitations of this study will be discussed. The AtomBench benchmarking framework is not a new benchmarking platform; rather, it corresponds to a new class of entries on the JARVIS-Leaderboard, an extensive, community-driven AI-for-materials benchmarking platform with hundreds of other entries. For this study, we explicitly benchmark the crystal reconstruction task applied to superconducting crystals. This benchmark is not exhaustive, as we did not consider other generative crystal reconstruction models in this work, such as DiffCSP, SymmCD, and CrystaLLM-$\pi$. However, AtomBench is designed to be modular and extensible, enabling future benchmarking of these models. We also clarify that we are measuring the models' ability to predict crystal structures using precalculated T$_c$ labels; evaluating the accuracy of these T$_c$ labels is outside the scope of this project. The focus of this benchmark is reconstruction fidelity, with T$_c$ as a conditioning variable, rather than end-to-end validation of generated crystals.

Next, differences in the datasets used in this study will be described. Given that generative crystal models inherit the biases and chemical-validity assumptions of the DFT methodology used to generate their training and testing sets, understanding the differences between the JARVIS Supercon-3D and Alexandria DS-A/B datasets~\cite{cerqueira2024sampling, schmidt2024improving} provides important context for this study's results. The first dataset we use is the JARVIS Supercon-3D superconductor dataset~\cite{Choudhary_2022}, and the second dataset is the Alexandria DS-A/B dataset~\cite{cerqueira2024sampling}. The crystal/T$_c$ pairs in both datasets were generated using Density Functional Perturbation Theory (DFPT), but with different protocols and screening philosophies.

\begin{figure}[ht]
  \centering
  \begin{subfigure}[t]{0.35\textwidth}
    \centering
    \includegraphics[width=\linewidth]{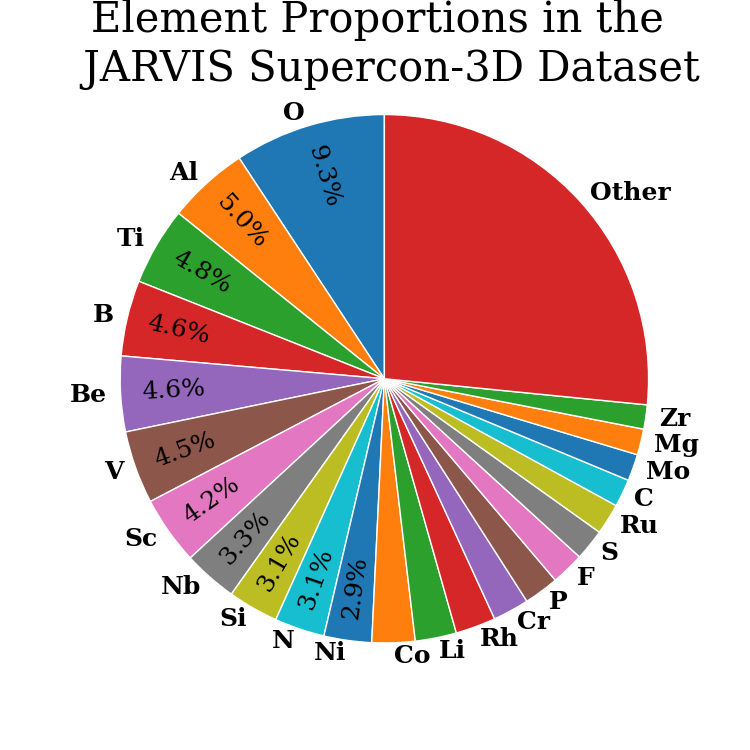}
    \subcaption{}
    \label{fig:jarvis-comp}
  \end{subfigure}
  \hspace{1em}
  \begin{subfigure}[t]{0.35\textwidth}
    \centering
    \includegraphics[width=\linewidth]{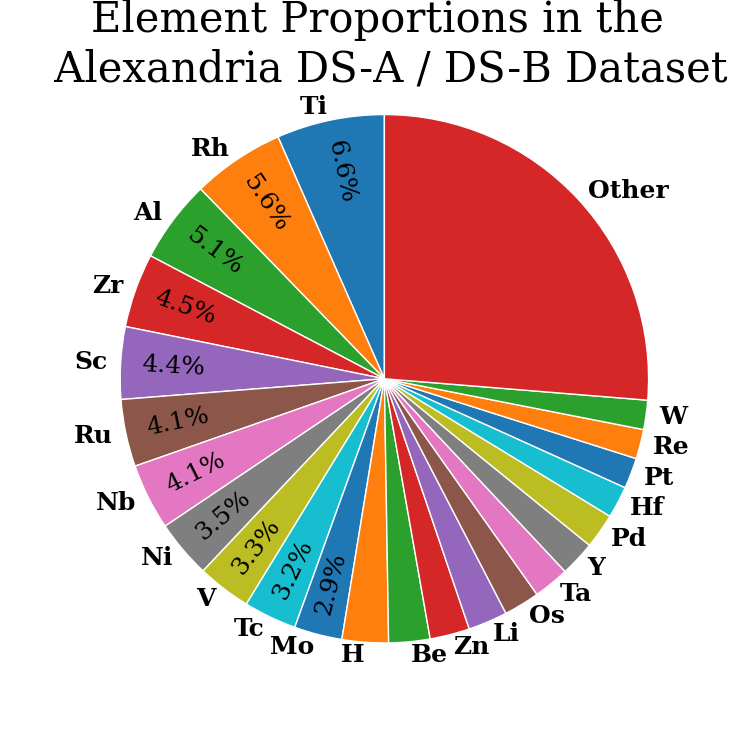}
    \subcaption{}
    \label{fig:alex-comp}
  \end{subfigure}

  \caption{
  Elemental composition of the superconductivity datasets used in this study.
  Pie charts show the 23 most frequently occurring elements in
  (a) JARVIS Supercon-3D and (b) Alexandria DS-A/B.
  JARVIS Supercon-3D is dominated by oxygen, aluminum, and titanium
  (9.3\%, 5.0\%, and 4.8\%, respectively),
  while Alexandria DS-A/B exhibits a higher prevalence of titanium,
  rhodium, and aluminum (6.6\%, 5.0\%, and 5.1\%).
  }
  \label{fig:dataset-composition}
\end{figure}
 
In JARVIS Supercon-3D, there are a total of 1,058 crystal/T$_c$ pairs. The Morel-Anderson effective Coulomb pseudopotential is set to $\mu^*=0.09$, the PBEsol exchange correlation functional is used, GBRV pseudopotentials are used, and convergence is confirmed if DFPT phonon spectra exhibit no imaginary phonon modes at any of the sampled Brillouin zone q-points. From Figure 3, the top 5 most represented elements in JARVIS Supercon-3D are oxygen, aluminum, titanium, boron, and beryllium, respectively, and most of the materials fall into the cubic, tetragonal, and hexagonal crystal systems.

In Alexandria DS-A/B, there are a total of 8,253 crystal/T$_c$ pairs. The Morel-Anderson effective Coulomb pseudopotential is set to $\mu^*=0.10$, the PBEsol exchange correlation functional is used, PseudoDojo pseudopotentials are used, and convergence is confirmed if DFPT phonon spectra exhibit a small number of imaginary phonon modes (under a small threshold) at any of the sampled Brillouin zone q-points. From Figure 3, the top 5 most-represented elements in Alexandria DS-A/B are titanium, rhodium, aluminum, zirconium, and scandium, respectively, and most of the structures belong to the tetragonal and cubic crystal systems.

\begin{figure}[ht]
  \centering
  \begin{subfigure}[t]{0.35\textwidth}
    \centering
    \raisebox{0.65em}{
      \includegraphics[width=\linewidth]{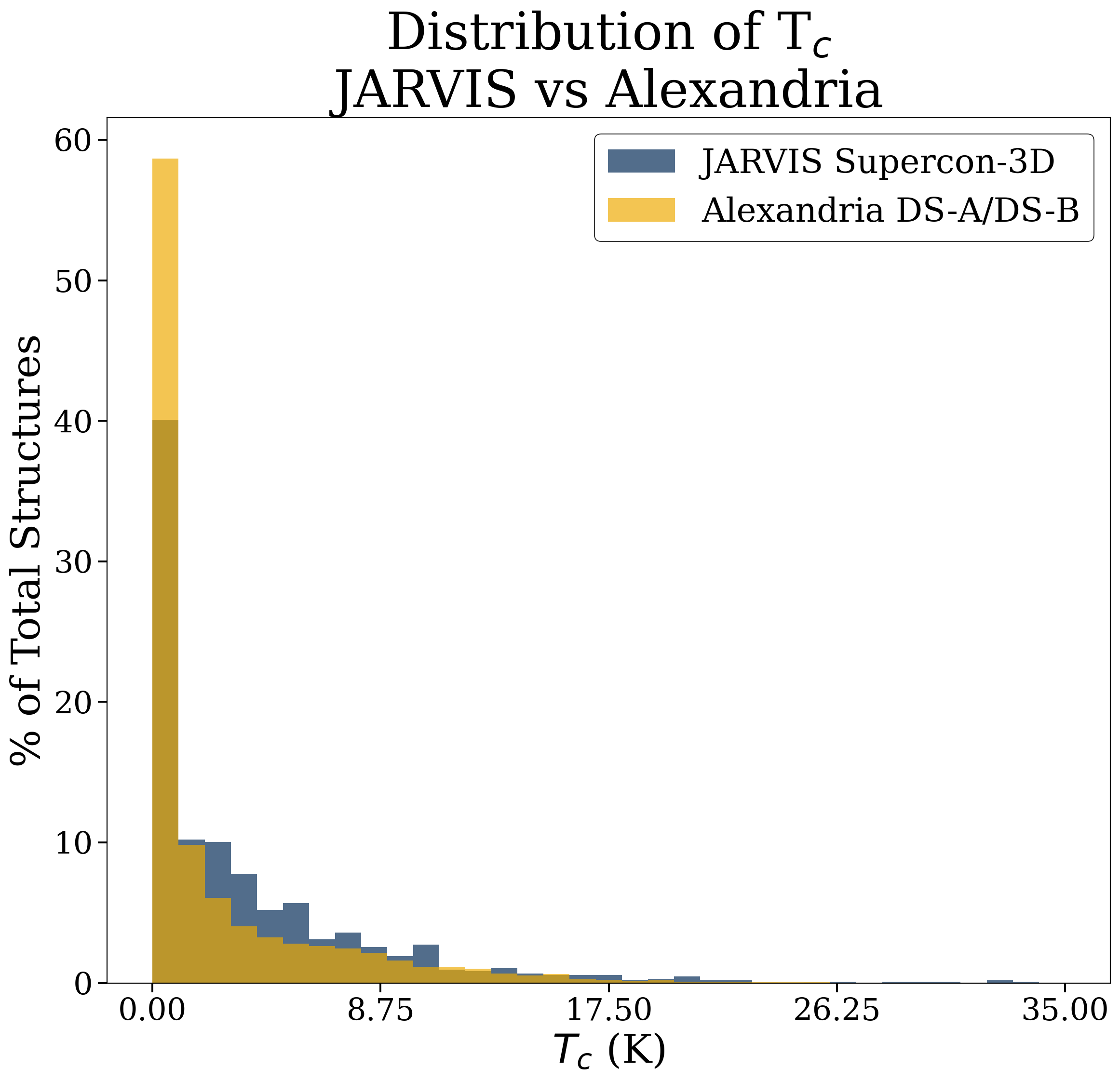}
    }
    \subcaption{}
    \label{fig:tc-overlay}
  \end{subfigure}
  \hspace{1em}
  \begin{subfigure}[t]{0.35\textwidth}
    \centering
    \includegraphics[width=\linewidth]{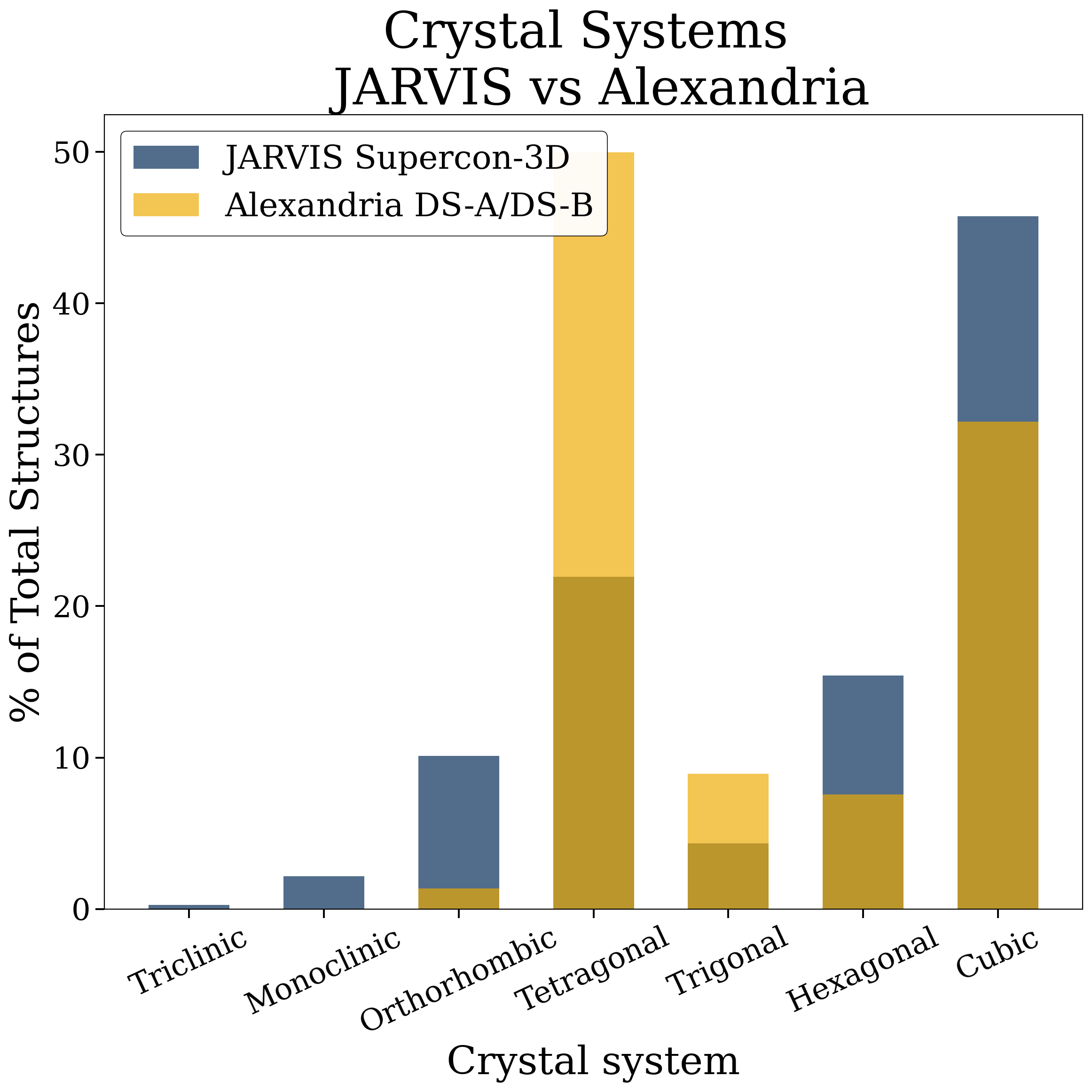}
    \subcaption{}
    \label{fig:crystal-overlay}
  \end{subfigure}

  \caption{
  Statistical comparison of superconducting and crystallographic properties
  between JARVIS Supercon-3D and Alexandria DS-A/B.
  (a) Overlaid histograms of superconducting critical temperature ($T_c$)
  show approximately exponential decay in both datasets, with
  Supercon-3D containing a larger fraction of high-$T_c$ materials.
  (b) Overlain bar charts of crystal system distributions indicate that
  Alexandria DS-A/B is dominated by cubic and tetragonal structures
  ($>$80\%), whereas Supercon-3D exhibits greater crystallographic diversity,
  with $\sim$45\% cubic phases and non-negligible representation across all
  seven crystal systems.
  }
  \label{fig:dataset-statistics}
\end{figure}
 
For both datasets, crystal systems were determined using pymatgen’s SpacegroupAnalyzer with a symmetry tolerance of 0.1 Å and an angle tolerance of 5°, and both datasets are dominated by low-T$_c$ entries with a sparse, high-T$_c$ tail. More information about protocols and screening can be found in the appendix.

An important note is that T$_c$ is used only as a conditioning variable for crystal prediction, and we do not interpret absolute T$_c$ as directly comparable between different datasets.

We first define the relationship between two sets of crystals: (1) the 10\% subset of fully specified test set crystals $\mathbf{M_i}$ and (2) the corresponding collection of crystals whose geometries are predicted by the inverse models $\hat{\mathbf{M}}_i$. Each crystal in the test set is fully specified, meaning that its chemical species $\mathbf{A}$, lattice parameters $\mathbf{L}$, atomic coordinates $\mathbf{X}$, and superconducting transition temperature T$_c$ are known. For each test set crystal, there is a corresponding partially-specified crystal that shares the same values of $\mathbf{A}$ and T$_c$, and importantly, the partially-specified crystal does not contain any values of $\mathbf{L}$ or $\mathbf{X}$. It is the purpose of the inverse model to map the partially-specified crystal's $\mathbf{A}$ and T$_c$ to a new $\hat{\mathbf{L}}$ and $\hat{\mathbf{X}}$ as a function of the inverse model’s learned parameters. We denote the set of partially-specified crystals that have undergone this mapping to be the set of predicted crystals, or, interchangeably, the set of reconstructed crystals. In the CDVAE setting, test structures $\mathbf{M_i}$ are encoded into latent representations $\mathbf{z}_i$ and
decoded back into test structures $\hat{\mathbf{M}}_i$. For consistency with the above protocol, the decoded test structures $\hat{\mathbf{M}}_i$ are treated as reconstructions of $\mathbf{M_i}$ and are evaluated with the same metrics; model-specific
details appear in the CDVAE section.

Second, we specify data leakage, disjoint train/test sets, and cross-validation considerations. For this benchmark, some models use a validation set, while others do not. For models that use a validation set, we employ an 80-10-10 train/test/validation split. For models that do not use a validation set, we employ a 90-10 train/test split, where the training split is constructed by concatenating the training and validation sets used by models with a validation set. In all cases, identical test sets are used across benchmarks. We ensure disjoint train and test partitions, zero data leakage, and structural uniqueness using CIF-text hashing, and we verify that no structural, lattice, or coordinate information from the test set leaks into the training or validation sets. The datasets were deterministically shuffled using base Python with a fixed random seed (123). Finally, cross-validation was intentionally omitted from this study, as it would introduce fold-dependent variation that could confound direct comparisons of relative model performance.

Before detailing the statistical assessment, we must first define how a crystal structure is parameterized. All information about a bulk crystal can be derived from its unit cell, the fundamental repeating building block that generates the crystal through periodic translation. The unit cell has the shape of a parallelepiped, and the lengths and angles of the edges are given by six lattice parameters $\{a, b, c, \alpha, \beta, \gamma\}$. Atoms reside in this repeated parallelepiped, and their positions are given by the atomic position vectors ${\mathbf{r}_i}$. Together, these parameters fully specify the periodic arrangement of atoms used as input for both training and reconstruction analysis. For more information about how bulk crystals are mathematically defined, see Ref.\cite{ashcroft_mermin_ssp}.

 Prior to computing statistical metrics, the Niggli reduction is applied to all reconstructed structures, as this forces the ordering of lattice constants and angles to be dependent on the geometry of the unit cell, rather than on choice or labeling convention\cite{niggli}. This greatly reduces ambiguity in lattice parameter labeling and makes statistical analysis of lattice parameter predictions meaningful.

For the first statistical metric, we have employed the Mean Absolute Error (MAE) to quantify the average deviation between predicted and ground-truth lattice parameters. Specifically, the MAE is computed for each of the six lattice parameters $\{a, b, c, \alpha, \beta, \gamma\}$ by averaging the absolute differences between their predicted values $\hat{y}_i$ and reference values $y_i$, $\mathrm{MAE} \;=\; \frac{1}{n} \sum_{i=1}^{n} \left| y_i - \hat{y}_i \right|$. Consequently, for each trained model instance, six Mean Absolute Error (MAE) values are produced. These values directly indicate the model's precision in reproducing the crystal geometry. An accurate reproduction of the lattice is essential for determining the correct space group and the crystal family. The MAE is a standard and interpretable metric widely used to assess the average magnitude of prediction errors.

We also use Kullback-Leibler Divergence (KLD)\cite{KLD} as a directed divergence between the histogrammed distribution of a single ground-truth crystal lattice parameter and the histogrammed distribution of a single predicted crystal lattice parameter for all six crystal lattice parameters $\{a,b,c,\alpha,\beta,\gamma\}$. KLD does not satisfy the formal axioms of a metric, but we still use its identity as a directed divergence to extract meaningful information about the predicted distributions. By convention, for a true distribution $P$ and a predicted distribution $Q$, the KLD between them quantifies the expected log‐ratio of probabilities of events $p(x)$ and $q(x)$ under distributions $P$ and $Q$ respectively:

\begin{equation}
D_{\mathrm{KL}}(P\|Q) \;=\; \sum_{x} p(x)\,\log\!\frac{p(x)}{q(x)}
\end{equation}

This value is widely interpreted as the amount of information lost when the distribution $Q$ approximates the distribution $P$, and we adopt the same interpretation in this study. A deeper analysis of Kullback-Leibler Divergence can be found in the original paper, "On Information and Sufficiency" \cite{KLD} by Kullback and Leibler.

 For the second statistical metric, we employ the average matched root-mean-squared displacement (RMSD) in Cartesian coordinates between the ground-truth and reconstructed sets. The RMSD between predicted and reference atomic positions $\hat{\mathbf{r}}_i$ and $\mathbf{r}_i$ is computed as $\mathrm{RMSD} = \sqrt{\frac{1}{N} \sum_{i=1}^{N} \|\mathbf{r}_i - \hat{\mathbf{r}}_i\|^2}$ and then averaged over all matched materials in the held-out test subset. Unlike the earlier metrics, only one mean RMSD is produced for each trained model instance. Because interatomic distances vary substantially across materials, we normalize the mean RMSD by an appropriate structural length scale to ensure comparability across datasets. Lower normalized mean RMSD values indicate that a model more faithfully reproduces the spatial arrangement of atoms within each crystal, providing a direct measure of local structural accuracy that complements the global lattice and distributional metrics. We use pymatgen StructureMatcher with STOL=0.5 to compute RMSD values. StructureMatcher compares atomic coordinates relative to the position of a fixed atom and accounts for lattice transformations, translations, and atomic permutations prior to RMSD evaluation.

{The mean RMSD characterizes only the structures that the StructureMatcher accepts within its tolerance, and it is therefore undefined in practice for models that match few or no structures. 
As a fourth statistical metric, we introduce a continuous, match-rate-independent measure of reconstruction fidelity, which we term the continuous corrected RMSD (ccRMSD). This metric is based on the average minimum distance (AMD), a unit-cell-independent isometry invariant of periodic crystals that varies continuously under perturbations of the atomic coordinates~\cite{widdowson2022amd}. AMD can be viewed as the average of the pointwise distance distribution (PDD), which characterizes the local environment of each atom through the ordered distances to its neighboring atoms.
For a crystal $\mathbf{M}$ whose unit cell contains m atoms, the order-k average minimum distance is the vector $\mathrm{AMD}_k(\textbf M)\in\mathbb{R}^{k}$, whose $i$-th component is the average distance from each atom in the unit cell to its $i$-th nearest neighbor in the infinite periodic structure:}

\begin{equation}
{\mathrm{AMD}_k(\textbf M)_i \;=\; \frac{1}{m}\sum_{j=1}^{m} d_{ij},}
\end{equation}

{
where $d_{ij}$ denotes the distance from atom j to its $i$-th nearest neighbor. Because AMD is Lipschitz-continuous with respect to perturbations of the atomic coordinates, small geometric changes produce proportionally small changes in the descriptor~\cite{widdowson2022amd}. Consequently, geometrically similar crystal structures are represented by nearby AMD vectors.
We define ccRMSD over a held-out test set of N target/prediction pairs $\{(\mathbf M_n,\hat{\mathbf M}_n)\}_{n=1}^{N}$ as the root mean square, taken across the entire dataset, of the Chebyshev $(L_\infty)$ distance between the AMD descriptors of each target crystal and its reconstruction:}

\begin{equation}
{\mathrm{ccRMSD} \;=\; \sqrt{\frac{1}{N}\sum_{n=1}^{N}\big\lVert \mathrm{AMD}_k(\textbf M_n) - \mathrm{AMD}_k(\hat{\textbf M}_n)\big\rVert_{\infty}^{2}},}
\end{equation}

{where $\lVert \mathbf{u}\rVert_{\infty} = \max_i |u_i|$. We use the Chebyshev norm because it captures the largest discrepancy among the neighbor shells represented in the AMD descriptor, making the metric sensitive to localized reconstruction errors. Throughout this work, we use $k = 100$ nearest neighbors.
We refer to this quantity as a continuous corrected RMSD because it provides a continuous analogue of RMSD that remains well defined even when exact atomic matching fails. Unlike the matched RMSD, ccRMSD does not rely on a match/no-match threshold and is evaluated for every structure in the test set, regardless of whether a successful structural match is obtained. As a result, the metric remains meaningful even when the match rate approaches zero.
Because both the AMD descriptor and its pairwise differences have units of length, ccRMSD is naturally reported in \AA{}, facilitating comparison with conventional RMSD values. Thus, ccRMSD measures the fidelity of the reconstructed local atomic geometry, while the locality of the metric is controlled by the hyperparameter k, which determines the number of nearest-neighbor shells included in the descriptor.}

Although each of the above metrics provides a quantitative measure of reconstruction fidelity, they probe distinct and inherently incomplete aspects of crystal structure similarity. Consequently, no single metric fully characterizes reconstruction quality, and each must be interpreted in light of the structural information it emphasizes and suppresses. 

The lattice‐parameter MAE directly measures the per‐structure geometric accuracy of the reconstructed unit cell. It is sensitive to systematic distortions of the metric tensor, such as uniform expansion or shear misalignment, and therefore reflects the model’s ability to reproduce the global crystal geometry required for correct symmetry classification. However, MAE is entirely insensitive to the statistical structure of the predicted ensemble: a model may achieve a low MAE while producing lattice parameters whose population‐level distribution deviates substantially from that of the ground‐truth dataset. Moreover, MAE does not encode any information about atomic positions within the unit cell and therefore cannot detect local structural errors once the lattice is fixed. 

In contrast, the Kullback-Leibler divergence probes agreement at the level of distributions rather than individual structures. Low KLD values indicate that the predicted ensemble reproduces the empirical distribution of lattice parameters observed in the reference dataset. However, KLD is agnostic to the correspondence between individual predicted structures and their ground‐truth counterparts, and as a result, substantial per‐structure errors may be concealed if they preserve the overall distribution. In this sense, KLD can mask large local deviations while still indicating high statistical fidelity, and should therefore not be interpreted as a measure of reconstruction accuracy on a per‐sample basis.

The normalized RMSD complements the lattice‐based metrics by directly assessing local atomic geometry. It is sensitive to errors in relative atomic positions and short‐range structural motifs that are invisible to lattice‐only measures. Consequently, RMSD provides a stringent test of whether reconstructed structures preserve chemically meaningful local environments. Nevertheless, RMSD does not encode global distributional information and cannot distinguish between systematic and stochastic errors across the dataset. Additionally, because RMSD is computed only after successful structural matching, it implicitly conditions on the existence of a reasonable correspondence between predicted and reference structures, thereby excluding certain classes of gross reconstruction failures from the statistic.

{The continuous corrected RMSD (ccRMSD) is designed to address precisely the blind spot just described. Because it is built on the Lipschitz-continuous AMD invariant and aggregates the $L_\infty$ distance between descriptors over the full test set, ccRMSD removes the match/no-match discontinuity inherent to the matched RMSD and is defined for every reconstructed structure, whether or not a structural correspondence is found. In this sense, it occupies a middle ground between the per-structure RMSD and the population-level KLD: like RMSD, it is computed pairwise between each target and its corresponding reconstruction. However, unlike matched RMSD, ccRMSD does not require or preserve an explicit sitewise correspondence between individual atoms; instead, it compares target and reconstructed crystals using averaged periodic nearest-neighbor distance descriptors, thereby measuring local periodic geometry rather than direct coordinate displacement. Like KLD, it is a single set-level statistic that summarizes the entire ensemble rather than a favorably matched subsample. This property is most valuable when a model matches only a fraction of the test set, where the matched RMSD can appear excellent while concealing systematic failure across the unmatched majority, whereas ccRMSD reflects the fidelity of the set of reconstructed crystals on a common, continuous scale.
}

Taken together, these metrics form a complementary set: MAE quantifies per‐structure global geometry, KLD captures population‐level statistical fidelity, and RMSD probes local atomic accuracy{, while ccRMSD provides a continuous, match-rate-independent measure of local geometric fidelity defined over every structure in the test set}. Interpreting them jointly mitigates the blind spots inherent to any individual metric and enables a more nuanced assessment of crystal reconstruction quality.

Next, we summarize the four model architectures, focusing on each model's paradigm and the information it receives about the target crystal prior to reconstruction; the most important aspects for this study are collected in Table~\ref{tab:architecture}. {Full architectural descriptions are deferred to the per-model appendices (Appendices~\ref{atomgpt}--\ref{mattergen}) so that readers can directly verify the differences in conditioning information that underlie the conditional-entropy regimes discussed later.}

\begin{table*}[t]
\centering
\small
\setlength{\tabcolsep}{4pt}
\renewcommand{\arraystretch}{1.15}
\caption{ {Architectural comparison of generative crystal models. Each of the four models benchmarked in this study is shown alongside the most important information about their architectures for the purposes of this study. When reconstructing a crystal structure, AtomGPT and MatterGen are the only models that can condition on the target crystal's T$_c$. Moreover, for large values of the CDVAE crystal embedding dimension, we find that there is an unequal amount of known information about the target crystal prior to reconstruction for each model, with CDVAE having the most information, followed by AtomGPT and MatterGen, and then FlowMM. {These conditioning regimes are summarized in Table~\ref{tab:regimes}.}}}
\label{tab:architecture}
\begin{tabularx}{\textwidth}{@{}l >{\raggedright\arraybackslash}X c >{\raggedright\arraybackslash}X >{\raggedright\arraybackslash}X@{}}
\hline
Model &
Backbone &
Conditions on T$_c$? &
Information Known Before Prediction &
Missing Information Predicted by the Model \\
\hline
FlowMM &
Riemannian Flow Matching &
No &
Atomic species &
Atomic coordinates, lattice parameters \\
\\
AtomGPT &
Generalist pretrained transformer finetuned on crystal/T$_c$ pairs &
Yes &
Atomic species, T$_c$ &
Atomic coordinates, lattice parameters \\
\\
CDVAE &
Diffusion VAE &
No &
Low-dimensional embedding of atomic species, atomic coordinates, and lattice parameters &
Atomic species, atomic coordinates, and lattice parameters \\
\\
MatterGen &
Equivariant joint diffusion (score network) with fine-tuned adapters &
Yes &
Atomic species, T$_c$ &
Atomic coordinates, lattice parameters \\
\hline
\end{tabularx}
\end{table*}

 AtomGPT is a generative pretrained transformer (GPT) that performs inverse design by representing crystals and target properties as text. A pretrained large language model (Mistral-7B) is finetuned on crystal/property pairs; at inference, it reads a prompt specifying the composition and, optionally, the target T$_c$, and generates the lattice parameters and atomic coordinates as a sequence of tokens. AtomGPT is thus conditioned on composition (and, when enabled, T$_c$) and predicts $\mathbf{L}$ and $\mathbf{X}$. The textual schema, finetuning procedure, and implementation details are given in the Appendix.

CDVAE is a diffusion variational autoencoder. For the reconstruction task, it encodes the full target structure $\mathbf{M}=(\mathbf{A},\mathbf{X},\mathbf{L})$ into a low-dimensional latent vector $\mathbf{z}$ using an SE(3)-equivariant periodic graph neural network, predicts aggregate properties (composition, lattice, and atom count) from $\mathbf{z}$, and then denoises a noised structure with a score-matching decoder to reconstruct $\mathbf{M}$. Because it conditions on a latent embedding of the \emph{entire} target crystal, CDVAE has access to more information about the target at inference than the other models (Table~\ref{tab:architecture}). The encoder/decoder architecture and the training loss are detailed in the Appendix.

FlowMM is a Riemannian flow-matching model. For crystal structure prediction, it fixes the composition $\mathbf{A}$ and learns a flow on the product manifold of fractional atomic coordinates and (Niggli-reduced) lattice parameters, integrating a base distribution forward in time to predict the atomic coordinates $\mathbf{X}$ and lattice $\mathbf{L}$. It is therefore conditioned on composition only. The manifold construction, the flow-matching objective, and the inference procedure are given in the Appendix.

MatterGen is a joint-diffusion model built on an SE(3)-equivariant score network and pre-trained on a large corpus of stable crystal structures. For crystal structure prediction, it holds the composition $\mathbf{A}$ fixed and denoises the atomic coordinates $\mathbf{X}$ and lattice $\mathbf{L}$ from noise; a target T$_c$ can optionally steer generation through fine-tuned adapter modules with classifier-free guidance. MatterGen is thus conditioned on composition (and, when enabled, T$_c$) and predicts $\mathbf{L}$ and $\mathbf{X}$. The diffusion processes, score network, and adapter-based conditioning are detailed in the Appendix.
\clearpage
\section{Results}
\label{sec:results}
\begin{figure}[hbt!]
\centering
\begin{subfigure}[t]{0.48\linewidth}
    \centering
    \includegraphics[width=\linewidth]{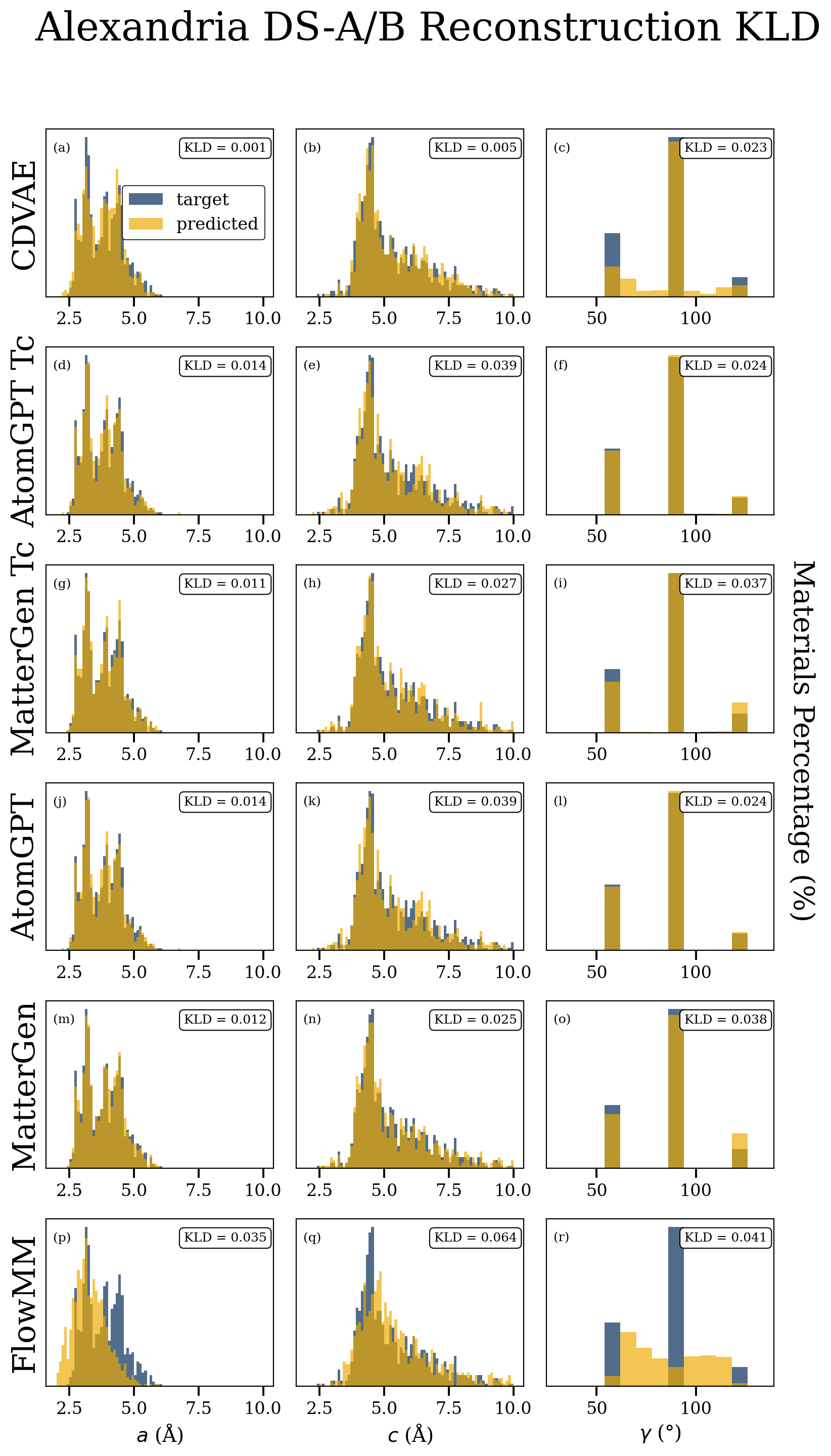}
    \caption{Alexandria DS-A/B test set.}
    \label{fig:dist_alexandria}
\end{subfigure}
\hfill
\begin{subfigure}[t]{0.493\linewidth}
    \centering
    \includegraphics[width=\linewidth]{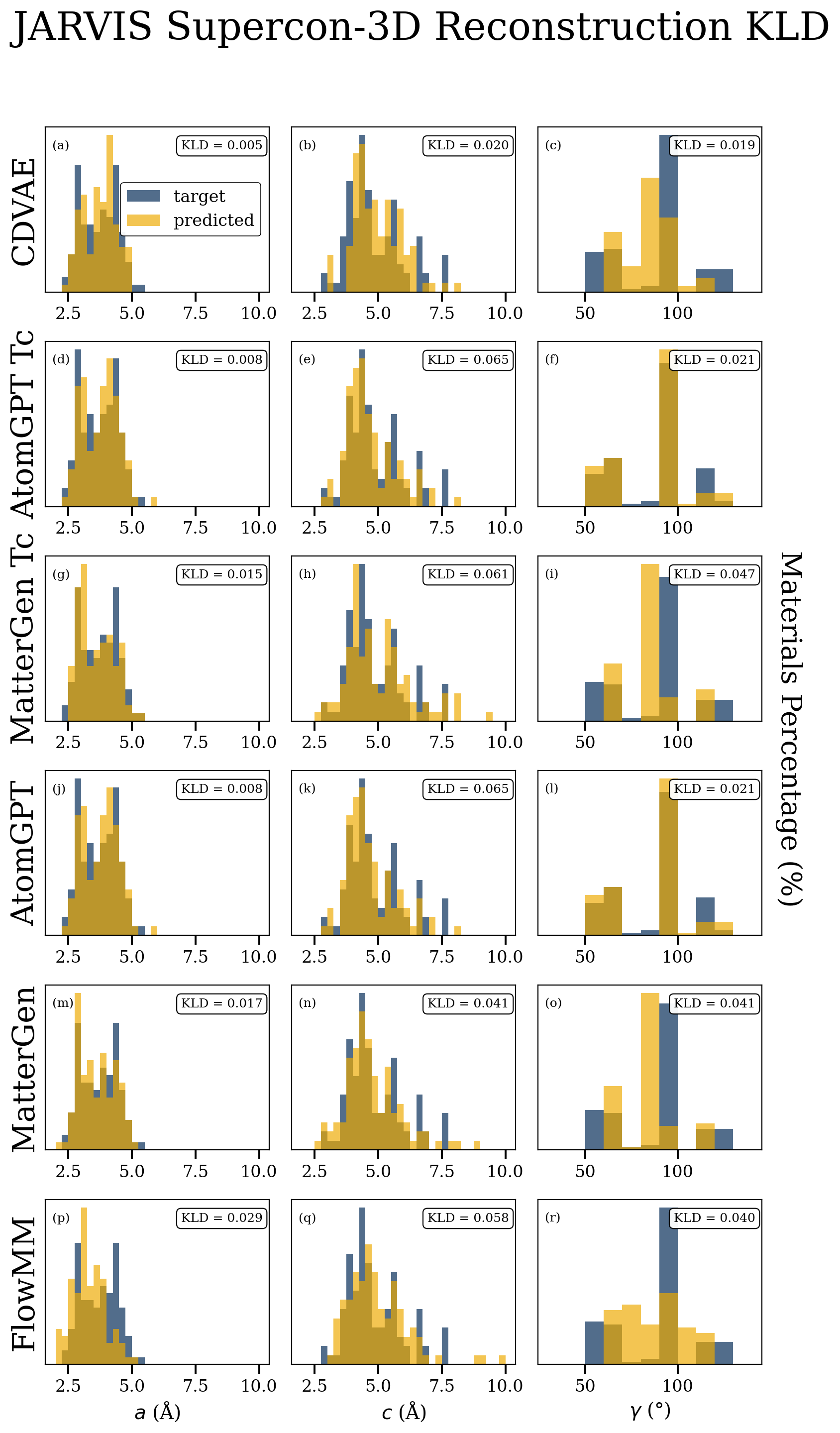}
    \caption{JARVIS Supercon-3D test set.}
    \label{fig:dist_jarvis}
\end{subfigure}

\caption{Reconstruction performance of CDVAE, AtomGPT, MatterGen, and FlowMM on the Alexandria DS-A/B and JARVIS Supercon-3D test sets for three representative Niggli-reduced lattice parameters, $a$, $c$, and $\gamma$. 'Tc' in the name string indicates that the model was conditioned on the target crystal's superconducting temperature during inference. The blue distributions are the target distributions directly obtained from the datasets, and the gold distributions are the predictions made by the models. For Alexandria and JARVIS, CDVAE, AtomGPT, and MatterGen perform similarly, with FlowMM matching least closely. Tc conditioning does not appear to improve lattice parameter distribution fidelity.}
\label{fig:reconstruction_grids}
\end{figure}
\clearpage
\begin{figure}[hbt!]
\centering

\begin{subfigure}[t]{0.46\linewidth}
    \centering
    \includegraphics[width=\linewidth]{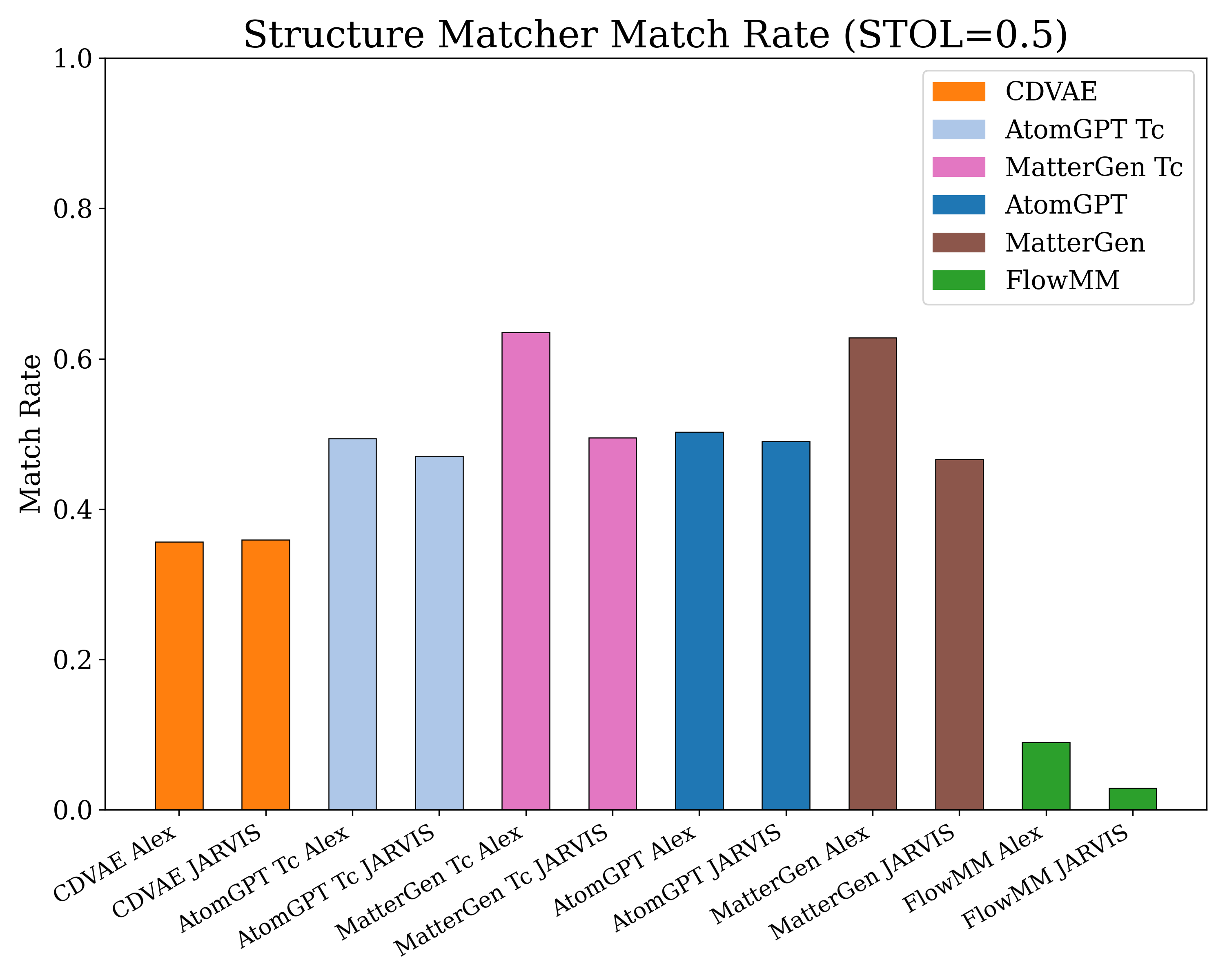}
    \caption{Reconstruction match rate across benchmark settings.}
    \label{fig:match_rate_bar_chart}
\end{subfigure}
\hfill
\begin{subfigure}[t]{0.49\linewidth}
    \centering
    \includegraphics[width=\linewidth]{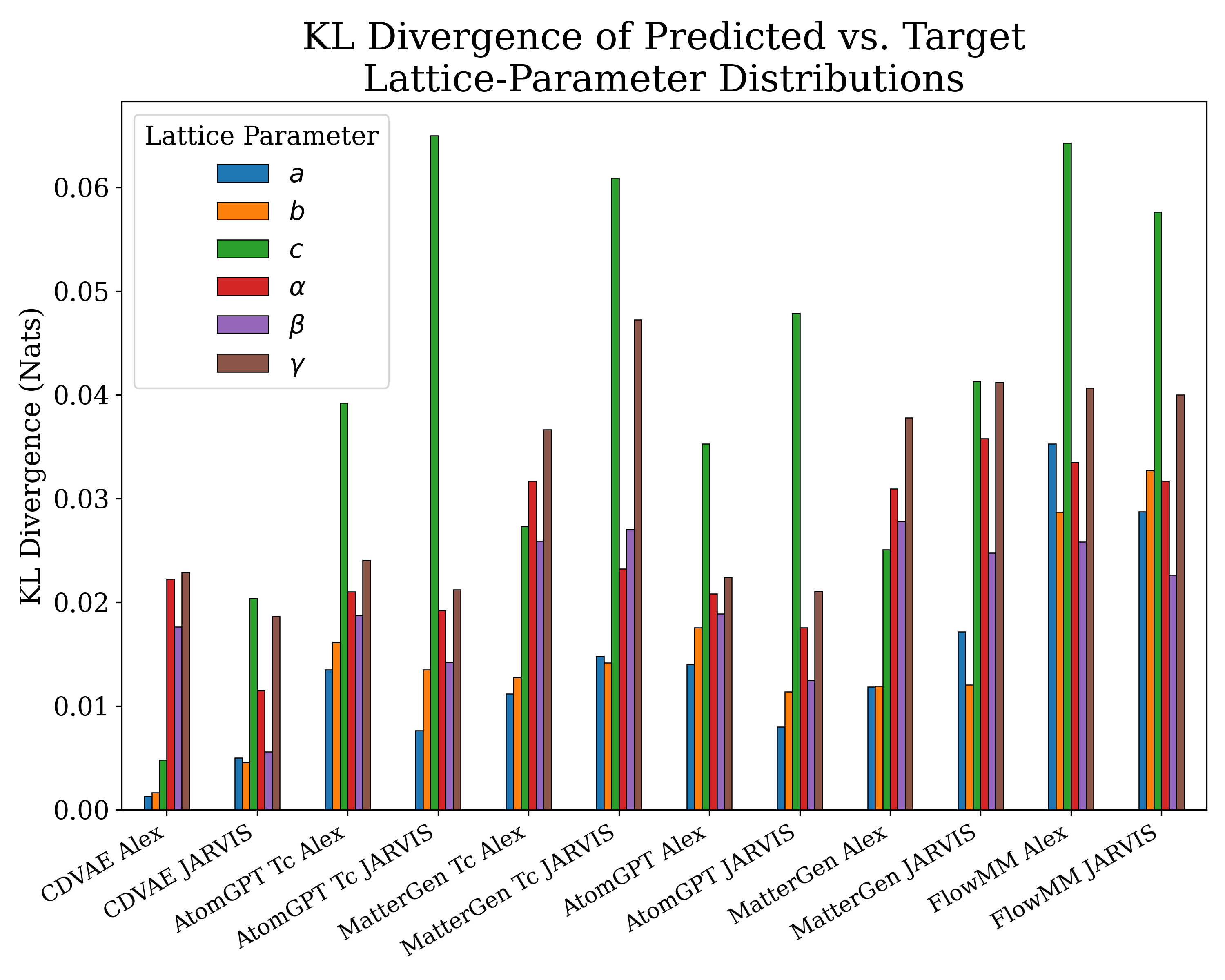}
    \caption{Kullback-Leibler divergence across all six Niggli-reduced lattice parameters.}
    \label{fig:kld_comparison_bar_chart}
\end{subfigure}

\caption{Comparison of reconstruction match rate and lattice parameter reconstruction KLD across benchmark settings. The match-rate panel reports reconstruction match rate across a variety of benchmarks (computed using \texttt{pymatgen StructureMatcher(STOL=0.5)}). MatterGen performs best, followed by AtomGPT, CDVAE, and FlowMM. The KLD panel reports Kullback-Leibler divergence in nats between the predicted and target distributions for all six Niggli-reduced lattice parameters, $a$, $b$, $c$, $\alpha$, $\beta$, and $\gamma$, across 12 experiments using 4 models and 2 datasets. For the KLD comparison, CDVAE has the most favorable scores for both datasets, followed by AtomGPT and MatterGen (which perform comparably) and then FlowMM. Tc conditioning does not appear to improve the match rate or the fidelity of lattice parameter reconstruction.}
\label{fig:match_rate_and_kld}
\end{figure}
\FloatBarrier

\begin{figure}[hbt!]
\centering

\begin{subfigure}[t]{0.48\linewidth}
    \centering
    \includegraphics[width=\linewidth]{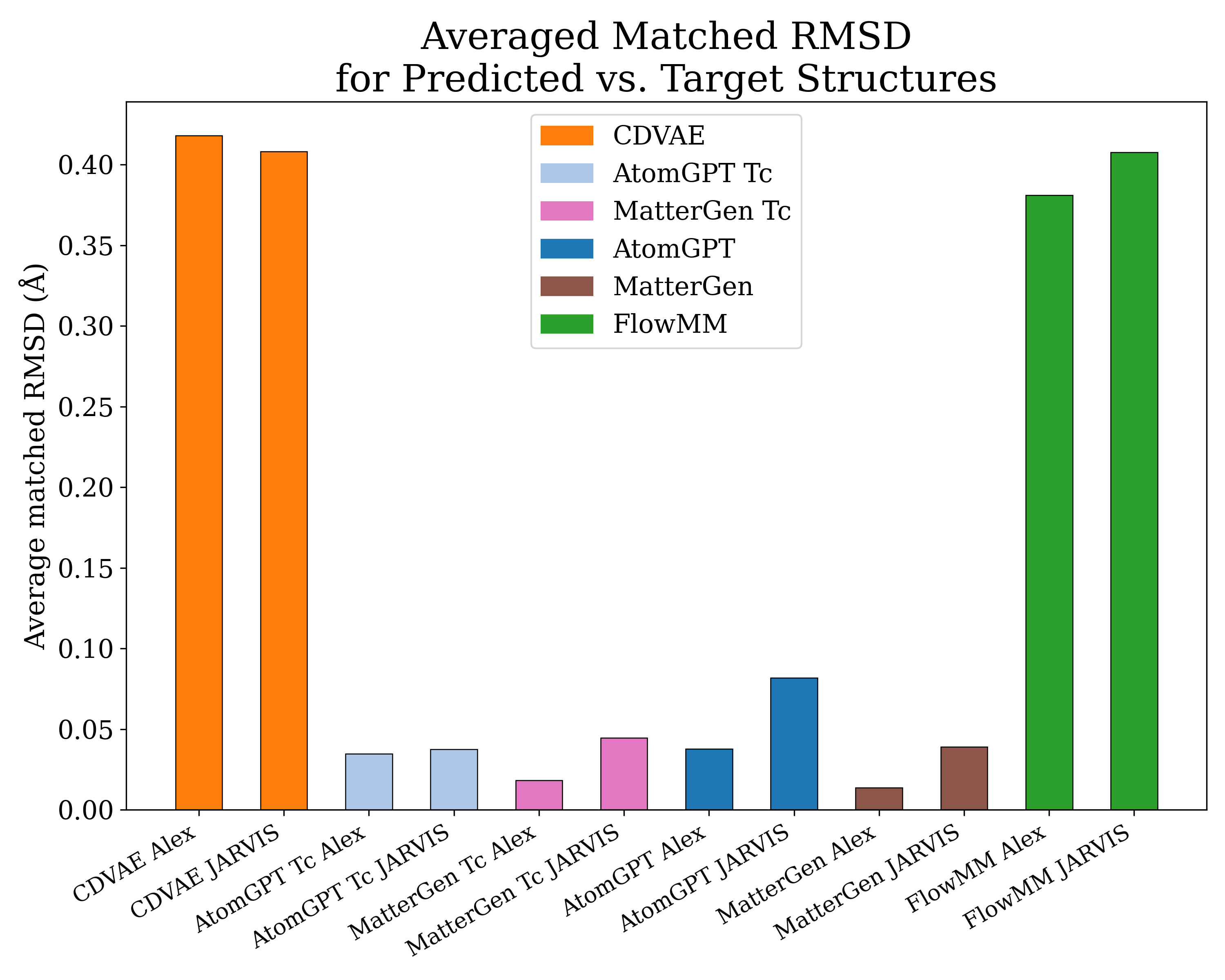}
    \caption{Average Matched RMSD.}
    \label{fig:cartesian_rmsd}
\end{subfigure}
\hfill
\begin{subfigure}[t]{0.48\linewidth}
    \centering
    \includegraphics[width=\linewidth]{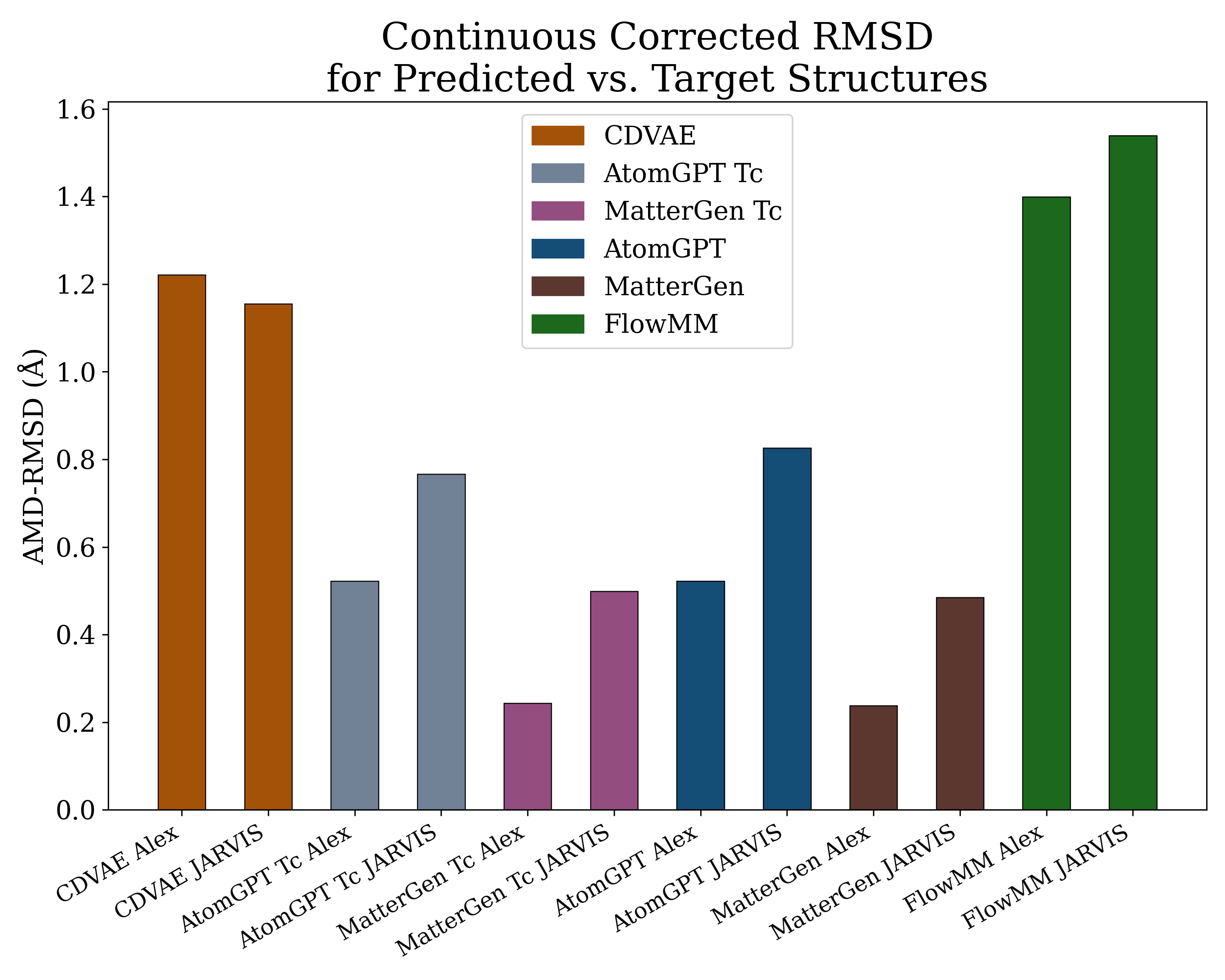}
    \caption{Continuous Corrected RMSD (ccRMSD).}
    \label{fig:ccrmsd}
\end{subfigure}

\caption{Atomic-coordinate reconstruction error between predicted and target structures for twelve experiments using four models and two datasets. Average Matched RMSD and continuous-corrected RMSD are both reported in angstroms. The ccRMSD metric is computed using k=100 nearest-neighbor atoms. MatterGen has the most favorable RMSD and ccRMSD across both datasets, followed by AtomGPT, while CDVAE and FlowMM perform similarly, with the highest error. Tc conditioning does not appear to improve the fidelity of atomic coordinate reconstruction.}
\label{fig:rmsd_bar_charts}
\end{figure}
\FloatBarrier

\begin{figure}[hbt!]
\centering

\begin{subfigure}[t]{0.48\linewidth}
    \centering
    \includegraphics[width=\linewidth]{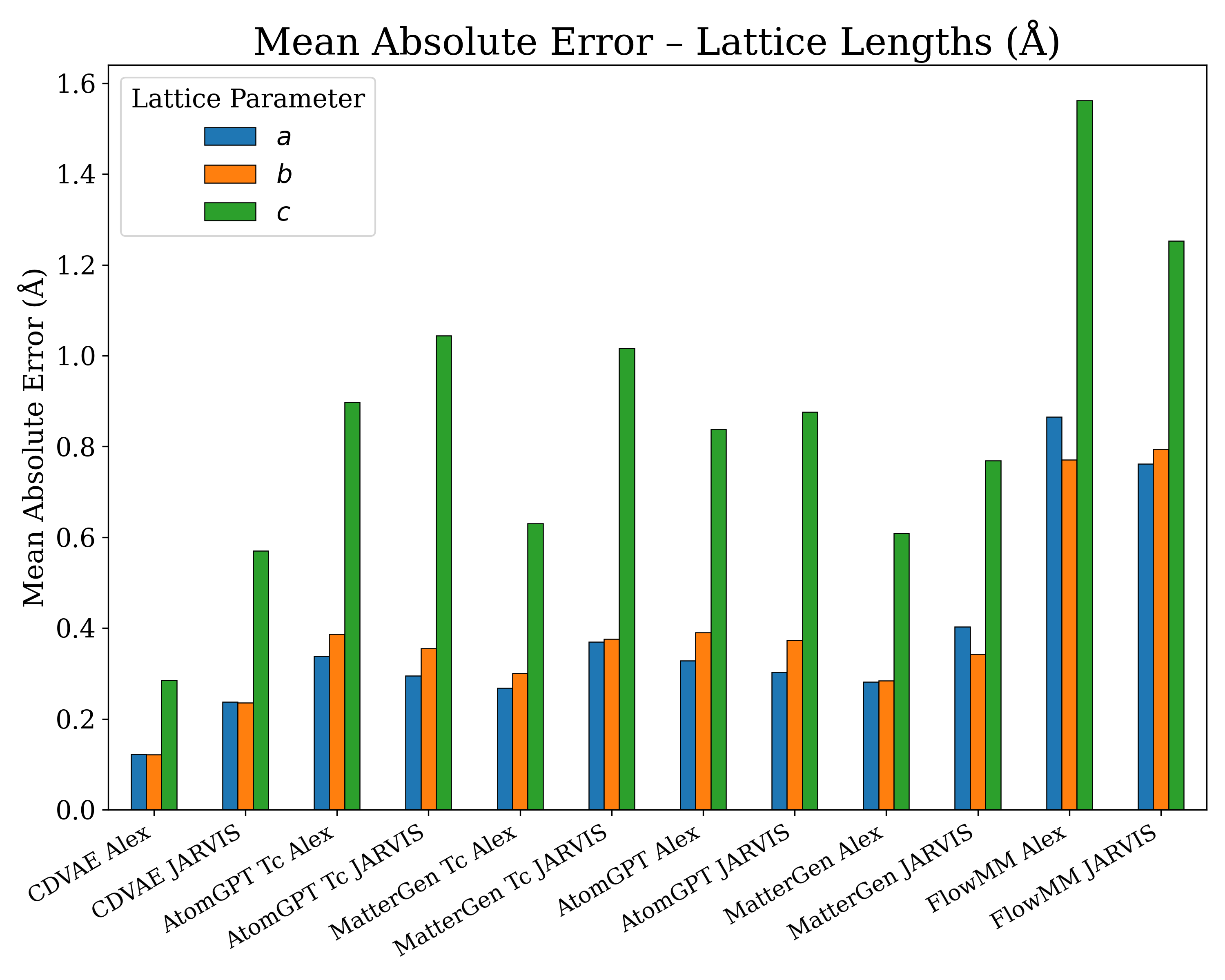}
    \caption{Niggli-reduced lattice lengths, $a$, $b$, and $c$.}
    \label{fig:mae_lengths}
\end{subfigure}
\hfill
\begin{subfigure}[t]{0.48\linewidth}
    \centering
    \includegraphics[width=\linewidth]{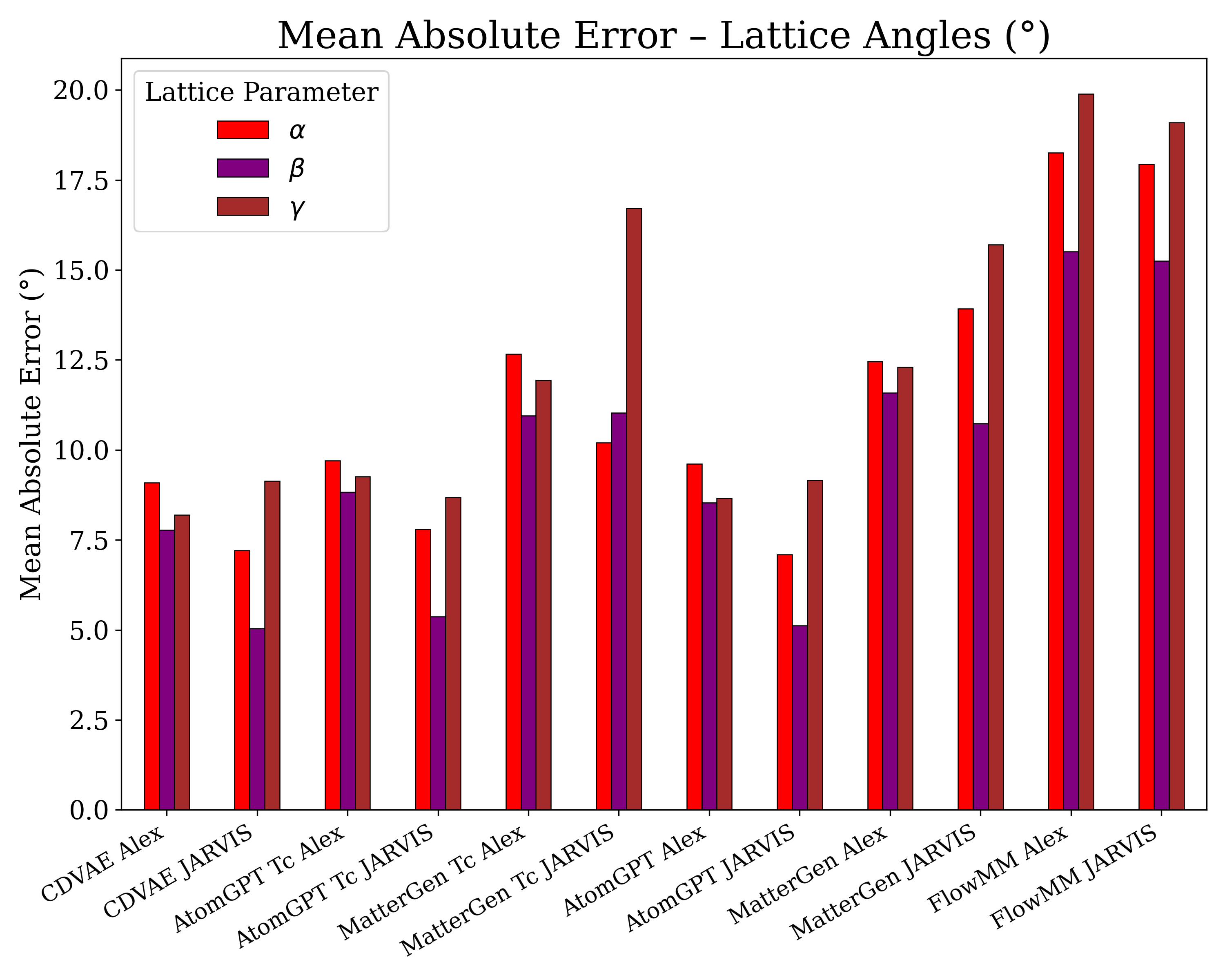}
    \caption{Niggli-reduced lattice angles, $\alpha$, $\beta$, and $\gamma$.}
    \label{fig:mae_angles}
\end{subfigure}

\caption{Mean absolute error between predicted and target Niggli-reduced lattice parameters for twelve experiments using four models and two datasets. Length errors are reported in angstroms for $a$, $b$, and $c$, while angle errors are reported in degrees for $\alpha$, $\beta$, and $\gamma$. CDVAE has the most favorable MAE scores for both datasets, followed by AtomGPT and MatterGen, and then FlowMM. Tc conditioning does not appear to improve lattice reconstruction MAE.}
\label{fig:mae_bar_charts}
\end{figure}
\FloatBarrier
\FloatBarrier

\begin{figure}[hbt!]
\centering

\begin{subfigure}[t]{0.48\linewidth}
    \centering
    \includegraphics[width=\linewidth]{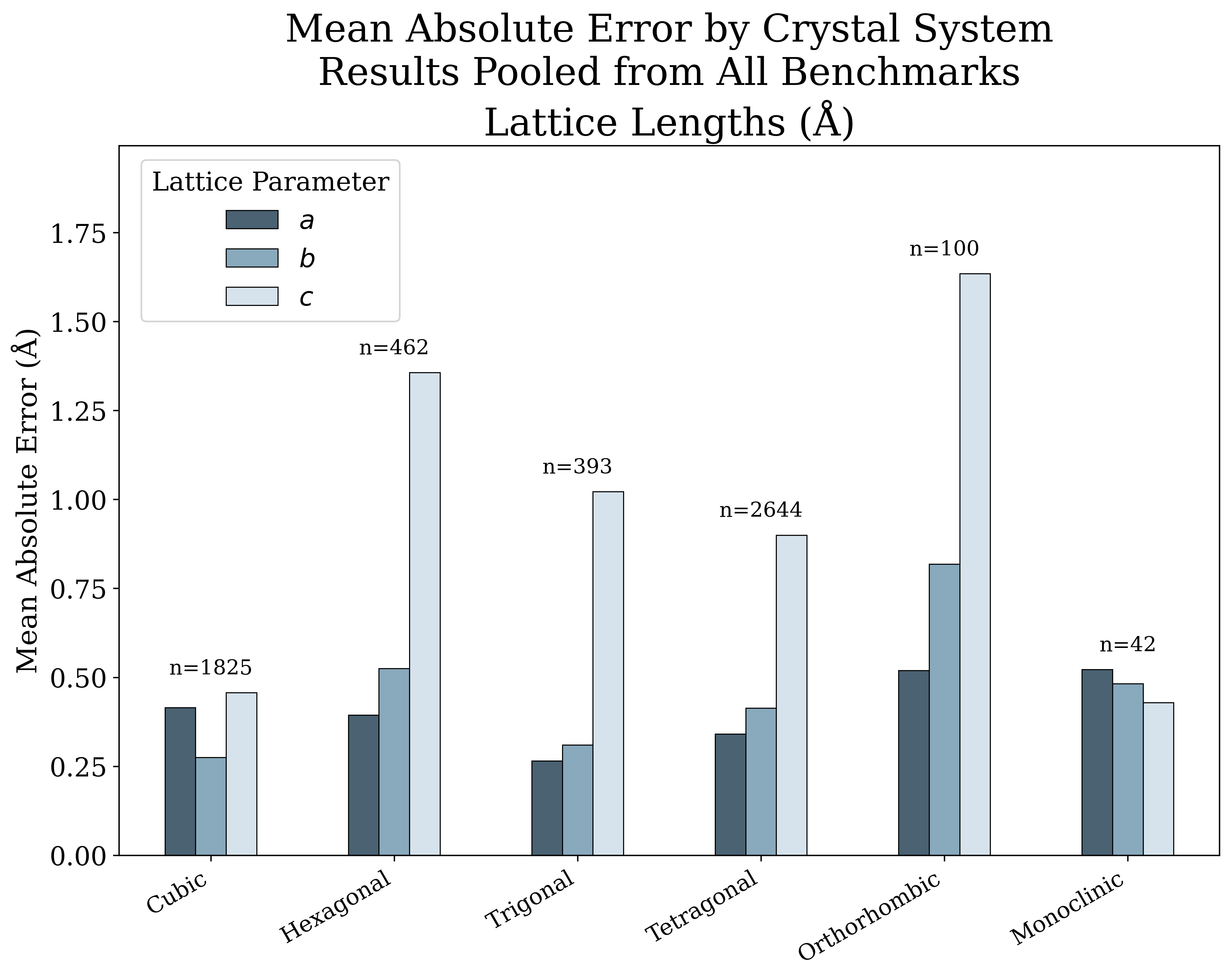}
    \caption{Mean lattice MAE in reconstructed lattice lengths, $a$, $b$, and $c$, by crystal system.}
    \label{fig:crystal_system_mae_lengths}
\end{subfigure}
\hfill
\begin{subfigure}[t]{0.48\linewidth}
    \centering
    \includegraphics[width=\linewidth]{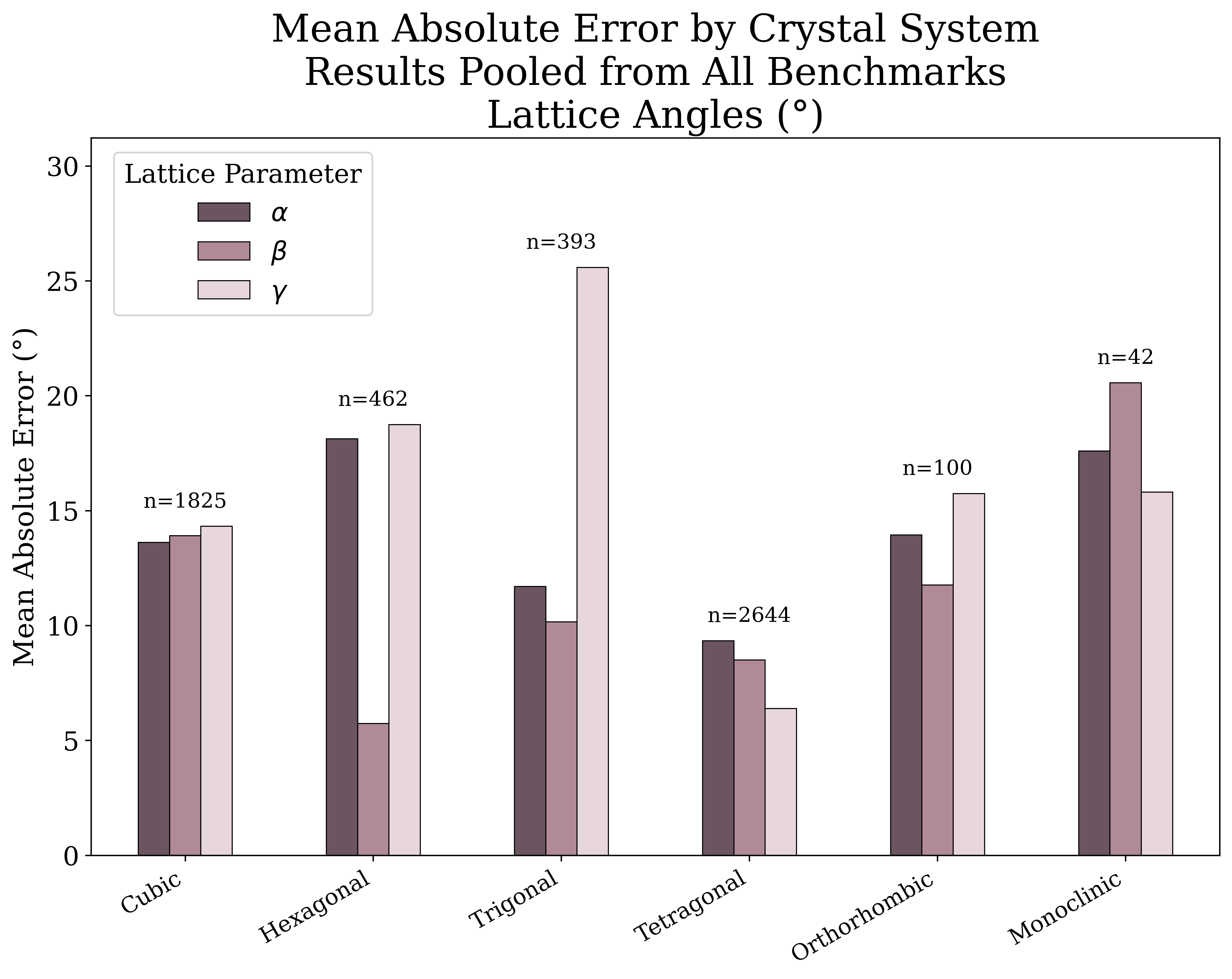}
    \caption{Mean lattice MAE in reconstructed lattice angles, $\alpha$, $\beta$, and $\gamma$, by crystal system.}
    \label{fig:crystal_system_mae_angles}
\end{subfigure}

\caption{Mean lattice MAE by crystal system, pooled across all twelve benchmark instances. Panels show MAE computed from Niggli-reduced primitive cells for lattice lengths ($a$, $b$, $c$) and lattice angles ($\alpha$, $\beta$, $\gamma$). Crystal-system labels are paired with their corresponding defining lattice constraints, and sample counts for each crystal system are shown above the grouped bars. Crystal families are ordered from most symmetric to least symmetric. Triclinic structures are omitted because fewer than 10 reconstructions belong to that family in this benchmark.}
\label{fig:crystal_system_mae}
\end{figure}

\begin{table*}[htbp]
\centering
\scriptsize

\begin{tabular}{llccccccc}
\multicolumn{9}{c}{\textbf{Lattice KLD}} \\
\hline
Dataset & Model & $a$ & $b$ & $c$ & $\alpha$ & $\beta$ & $\gamma$ & \textbf{mean} \\
\hline
 Alexandria & AtomGPT & 0.0140 & 0.0176 & 0.0353 & \textbf{0.0208} & 0.0189 & \textbf{0.0224} & 0.0215 \\
  & AtomGPT Tc & 0.0135 & 0.0162 & 0.0392 & 0.0210 & 0.0187 & 0.0240 & 0.0221 \\
  & CDVAE & \textbf{0.0013} & \textbf{0.0017} & \textbf{0.0048} & 0.0222 & \textbf{0.0176} & 0.0229 & \textbf{0.0118} \\
  & FlowMM & 0.0353 & 0.0287 & 0.0643 & 0.0335 & 0.0258 & 0.0407 & 0.0380 \\
  & MatterGen & 0.0119 & 0.0119 & 0.0251 & 0.0310 & 0.0278 & 0.0378 & 0.0242 \\
  & MatterGen Tc & 0.0112 & 0.0128 & 0.0273 & 0.0317 & 0.0259 & 0.0366 & 0.0243 \\
\hline
 JARVIS & AtomGPT & 0.0080 & 0.0114 & 0.0479 & 0.0176 & 0.0125 & 0.0211 & 0.0197 \\
  & AtomGPT Tc & 0.0077 & 0.0135 & 0.0650 & 0.0192 & 0.0142 & 0.0212 & 0.0235 \\
  & CDVAE & \textbf{0.0050} & \textbf{0.0046} & \textbf{0.0204} & \textbf{0.0115} & \textbf{0.0056} & \textbf{0.0187} & \textbf{0.0110} \\
  & FlowMM & 0.0288 & 0.0327 & 0.0576 & 0.0317 & 0.0226 & 0.0400 & 0.0356 \\
  & MatterGen & 0.0172 & 0.0121 & 0.0413 & 0.0358 & 0.0248 & 0.0412 & 0.0287 \\
  & MatterGen Tc & 0.0148 & 0.0142 & 0.0609 & 0.0232 & 0.0270 & 0.0472 & 0.0312 \\
\hline
\end{tabular}

\vspace{2em}

\begin{tabular}{llcccccccc}
\multicolumn{10}{c}{\textbf{Lattice MAE (\AA\ for $abc$, $^\circ$ for $\alpha\beta\gamma$)}} \\
\hline
Dataset & Model & $a$ & $b$ & $c$ & $\alpha$ & $\beta$ & $\gamma$ & \textbf{mean}$_{abc}$ & \textbf{mean}$_{\alpha\beta\gamma}$ \\
\hline
 Alexandria & AtomGPT & 0.329 & 0.390 & 0.839 & 9.612 & 8.539 & 8.659 & 0.519 & 8.937 \\
  & AtomGPT Tc & 0.338 & 0.387 & 0.898 & 9.700 & 8.828 & 9.259 & 0.541 & 9.262 \\
  & CDVAE & \textbf{0.123} & \textbf{0.122} & \textbf{0.285} & \textbf{9.087} & \textbf{7.776} & \textbf{8.190} & \textbf{0.177} & \textbf{8.351} \\
  & FlowMM & 0.865 & 0.770 & 1.562 & 18.251 & 15.512 & 19.886 & 1.066 & 17.883 \\
  & MatterGen & 0.282 & 0.284 & 0.609 & 12.460 & 11.584 & 12.299 & 0.392 & 12.114 \\
  & MatterGen Tc & 0.268 & 0.300 & 0.631 & 12.664 & 10.955 & 11.934 & 0.400 & 11.851 \\
\hline
 JARVIS & AtomGPT & 0.304 & 0.374 & 0.876 & \textbf{7.097} & 5.120 & 9.158 & 0.518 & \textbf{7.125} \\
  & AtomGPT Tc & 0.295 & 0.356 & 1.044 & 7.792 & 5.371 & \textbf{8.688} & 0.565 & 7.283 \\
  & CDVAE & \textbf{0.237} & \textbf{0.236} & \textbf{0.570} & 7.206 & \textbf{5.041} & 9.138 & \textbf{0.348} & 7.128 \\
  & FlowMM & 0.762 & 0.794 & 1.253 & 17.938 & 15.247 & 19.095 & 0.936 & 17.427 \\
  & MatterGen & 0.403 & 0.343 & 0.769 & 13.924 & 10.735 & 15.709 & 0.505 & 13.456 \\
  & MatterGen Tc & 0.369 & 0.376 & 1.016 & 10.201 & 11.034 & 16.710 & 0.587 & 12.648 \\
\hline
\end{tabular}

\vspace{2em}

\begin{tabular}{llc}
\multicolumn{3}{c}{\textbf{ccRMSD ($k=100$)}} \\
\hline
Dataset & Model & ccRMSD \\
\hline
 Alexandria & AtomGPT & 0.5224 \\
  & AtomGPT Tc & 0.5228 \\
  & CDVAE & 1.2218 \\
  & FlowMM & 1.3998 \\
  & MatterGen & \textbf{0.2380} \\
  & MatterGen Tc & 0.2440 \\
\hline
 JARVIS & AtomGPT & 0.8262 \\
  & AtomGPT Tc & 0.7665 \\
  & CDVAE & 1.1555 \\
  & FlowMM & 1.5392 \\
  & MatterGen & \textbf{0.4854} \\
  & MatterGen Tc & 0.4988 \\
\hline
\end{tabular}

\vspace{2em}

\begin{tabular}{llccc}
\multicolumn{5}{c}{\textbf{Average Matched RMSD}} \\
\hline
Dataset & Model & RMSD (\AA) & Match rate & STOL \\
\hline
 Alexandria & AtomGPT & 0.0378 & 0.5024 & 0.50 \\
  & AtomGPT Tc & 0.0347 & 0.4939 & 0.50 \\
  & CDVAE & 0.4181 & 0.3564 & 0.50 \\
  & FlowMM & 0.3810 & 0.0897 & 0.50 \\
  & MatterGen & \textbf{0.0138} & 0.6279 & 0.50 \\
  & MatterGen Tc & 0.0182 & \textbf{0.6352} & 0.50 \\
\hline
 JARVIS & AtomGPT & 0.0820 & 0.4902 & 0.50 \\
  & AtomGPT Tc & \textbf{0.0376} & 0.4706 & 0.50 \\
  & CDVAE & 0.4083 & 0.3592 & 0.50 \\
  & FlowMM & 0.4077 & 0.0291 & 0.50 \\
  & MatterGen & 0.0392 & 0.4660 & 0.50 \\
  & MatterGen Tc & 0.0446 & \textbf{0.4951} & 0.50 \\
\hline
\end{tabular}

\vspace{2em}

\caption{Reconstruction metrics for all benchmark instances. Stacked blocks: lattice KLD, lattice MAE (lengths in \AA, angles in degrees), matched RMSD, match rate, and STOL threshold, and ccRMSD (AMD-based RMSE, $k{=}100$ nearest neighbors). Bold marks the best value for each column in each dataset block. Match rate is bolded at its maximum; all other metrics at their minimum. The ccRMSD metric is computed using k=100 nearest-neighbor atoms.}
\label{tab:reconstruction_all}
\end{table*}

\begin{table*}[t]
\centering
\small
\setlength{\tabcolsep}{6pt}
\renewcommand{\arraystretch}{1.2}
\caption{{Conditioning-information regimes of the benchmarked models. Each benchmark is grouped by the information the model is given prior to reconstruction, following the categories in Table~\ref{tab:architecture}. Rows are ordered from the most to the least information known about the target crystal. The two models that support property conditioning (AtomGPT and MatterGen) each appear in two regimes: their composition-only variants (AtomGPT, MatterGen) share the composition-only regime with FlowMM, while their T$_c$-conditioned variants (AtomGPT Tc, MatterGen Tc) occupy the composition-plus-T$_c$ regime. CDVAE occupies a distinct regime in which a low-dimensional embedding of the entire target structure is available.}}
\label{tab:regimes}

\begin{tabularx}{\textwidth}{@{}l >{\raggedright\arraybackslash}X >{\raggedright\arraybackslash}X@{}}
\hline
Conditioning regime &
Information known before prediction &
Models \\
\hline
Full-structure embedding &
Low-dimensional embedding of atomic species, atomic coordinates, and lattice parameters $(\mathbf{A},\mathbf{X},\mathbf{L})$ &
CDVAE \\
\\
Composition $+$ T$_c$ &
Atomic species $\mathbf{A}$ and superconducting critical temperature T$_c$ &
AtomGPT Tc, MatterGen Tc \\
\\
Composition only &
Atomic species $\mathbf{A}$ &
FlowMM, AtomGPT, MatterGen \\
\hline
\end{tabularx}
\end{table*}

\begin{table*}[htbp]
\centering
\scriptsize

\begin{tabular}{lll}
\multicolumn{3}{c}{\textbf{Lattice Conditions Defining Crystal Systems}} \\
\hline
Crystal system & Length conditions & Angle conditions \\
\hline
Cubic        & $a=b=c$           & $\alpha=\beta=\gamma=90^\circ$ \\
Hexagonal    & $a=b\neq c$       & $\alpha=\beta=90^\circ,\ \gamma=120^\circ$ \\
Trigonal     & $a=b=c$           & $\alpha=\beta=\gamma\neq90^\circ$ \\
Tetragonal   & $a=b\neq c$       & $\alpha=\beta=\gamma=90^\circ$ \\
Orthorhombic & $a \neq b \neq c$ & $\alpha=\beta=\gamma=90^\circ$ \\
Monoclinic   & $a \neq b \neq c$ & $\alpha=\gamma=90^\circ,\ \beta\neq90^\circ$ \\
Triclinic    & $a \neq b \neq c$ & $\alpha \neq \beta \neq \gamma \neq 90^\circ$ \\
\hline
\end{tabular}

\vspace{2.6em}

\begin{tabular}{lccccccccc}
\multicolumn{10}{c}{\textbf{Aggregate Lattice MAE by Crystal System (Å for $abc$, $^\circ$ for $\alpha\beta\gamma$)}} \\
\hline
Crystal system & $n$
& $a$ & $b$ & $c$ & $\alpha$ & $\beta$ & $\gamma$ & mean$_{abc}$ & mean$_{\alpha\beta\gamma}$ \\
\hline
Cubic        & {1825} & {0.415} & {\textbf{0.275}} & {\textbf{0.457}} & {13.620} & {13.914} & {14.322} & {\textbf{0.382}} & {13.952} \\
Hexagonal    & {462} & {0.394} & {0.525} & {1.357} & {18.129} & {\textbf{5.743}} & {18.750} & {0.759} & {14.208} \\
Trigonal     & {447} & {\textbf{0.299}} & {0.342} & {1.079} & {11.568} & {9.951} & {26.038} & {0.573} & {15.852} \\
Tetragonal   & {2644} & {0.341} & {0.414} & {0.900} & {\textbf{9.343}} & {8.500} & {\textbf{6.394}} & {0.551} & {\textbf{8.079}} \\
Orthorhombic & {138} & {0.466} & {0.717} & {1.429} & {13.121} & {11.311} & {13.388} & {0.871} & {12.607} \\
Monoclinic   & {48} & {0.508} & {0.470} & {0.535} & {18.169} & {20.601} & {16.532} & {0.504} & {18.434} \\
Triclinic    & --   & --    & --    & --    & --     & --     & --     & --    & --     \\
\hline
\end{tabular}

\caption{Crystal-system-specific summary of reconstruction behavior pooled across all {twelve} benchmark instances. The top block lists the conventional lattice constraints that define the seven crystal families, ordered from most to least symmetric. The bottom block reports aggregate lattice mean absolute error by crystal system, with lengths in \AA\ and angles in degrees. Bold indicates the lowest value in each error column. Trigonal is shown in the rhombohedral setting. The triclinic family is included for completeness, but aggregate errors are not shown because fewer than 10 structures in this benchmark belong to that crystal family.}
\label{tab:crystal_system_summary}
\end{table*}

\begin{table*}[htbp]
\centering
\scriptsize

\begin{tabular}{lcccc}
\multicolumn{5}{c}{\shortstack{\textbf{Computational Cost Comparison of Generative Crystal Models}
}}  \\
\hline
Model & Total Parameters & Total Epochs & Time per Epoch & Time per Inference \\
\hline
FlowMM    & 14,200,000      & 10  & 12.9 seconds          & \textbf{0.5 seconds} \\
CDVAE     & 5,000,000       & 100 & 2.3 minutes           & 1.5 seconds \\
AtomGPT   & 7,000,000,000   & 10  & 15 minutes            & 4.6 seconds \\
MatterGen & 48,800,000      & 200 & \textbf{9.5 seconds}  & 5.3 seconds \\
\hline
\end{tabular}

\caption{Computational cost comparison of generative crystal models using Alexandria DS-A/B. Each of the four models benchmarked in this study is shown alongside computational cost information such as total parameters, total epochs, time per epoch, and time per inference. MatterGen has the most favorable training time per epoch, and FlowMM has the most favorable inference time. CDVAE and AtomGPT perform intermediately. MatterGen and AtomGPT utilize fine-tuning adapters over top pretrained foundation models, and CDVAE and FlowMM train from scratch. We use default hyperparameters for each model, and the difference in epochs reflects automatic early stopping.}
\label{tab:computationalcost}
\end{table*}

\newpage
In this reconstruction setting, across the Alexandria DS-A/B and JARVIS Supercon-3D benchmarks, CDVAE achieved the lowest reconstruction error for lattice parameter predictions measured by KLD and MAE. {AtomGPT and MatterGen performed comparably and intermediately between CDVAE and FlowMM, and FlowMM exhibited the highest lattice error. For atomic coordinate prediction, MatterGen achieved the lowest error, closely followed by AtomGPT, while CDVAE and FlowMM exhibited the highest coordinate error.}

{This atomic-coordinate ranking, however, depends on which metric is used. The matched RMSD (Figure~\ref{fig:cartesian_rmsd}) splits the four models into two stark regimes (MatterGen and AtomGPT with near-zero error, and CDVAE and FlowMM with large error), which would suggest that the two members of each pair perform comparably. The continuous corrected RMSD (Figure~\ref{fig:ccrmsd}; Table~\ref{tab:reconstruction_all}), which is evaluated over all structures rather than only matched ones, instead reveals a graded separation, $\mathrm{MatterGen} < \mathrm{AtomGPT} < \mathrm{CDVAE} < \mathrm{FlowMM}$ (Alexandria: $0.238 < 0.522 < 1.222 < 1.400$; JARVIS: $0.485 < 0.826 < 1.156 < 1.539$). This ordering mirrors the match-rate ranking of the models far more closely than the matched RMSD does, and the roughly even spacing between successive models indicates that AtomGPT and MatterGen are not in fact equivalent, nor are CDVAE and FlowMM. Critically, ccRMSD remains well-defined for FlowMM, whose sub-ten-percent match rate renders the matched RMSD statistically unreliable.}

{These reconstruction results can be organized by the information each model is given before prediction. As summarized in Table~\ref{tab:regimes}, and following the categories of Table~\ref{tab:architecture}, the twelve benchmarks fall into three conditioning-information regimes: a full-structure embedding regime (CDVAE); a composition-plus-T$_c$ regime occupied by the T$_c$-conditioned variants AtomGPT Tc and MatterGen Tc; and a composition-only regime occupied by FlowMM together with the composition-only variants AtomGPT and MatterGen. The composition-only regime is especially useful because it holds the conditioning information fixed across three architecturally distinct models. Within it, atomic-coordinate fidelity orders MatterGen $<$ AtomGPT $\ll$ FlowMM (ccRMSD $0.238 < 0.522 \ll 1.400$ on Alexandria and $0.485 < 0.826 \ll 1.539$ on JARVIS), and the match rate places MatterGen and AtomGPT comparably (MatterGen highest on Alexandria at $0.628$ versus $0.502$, and comparable on JARVIS at $0.466$ versus $0.490$), with both far above FlowMM ($0.090$ and $0.029$). Conditioning additionally on T$_c$ does not move either AtomGPT or MatterGen cleanly out of this ordering relative to its composition-only counterpart. These conditioning regimes are interpreted in the Discussion.}

Interestingly, each model tends to have a significantly higher relative error for the $c$ lattice parameter, with $c$ typically exhibiting errors approximately twice those of $a$ and $b$. Under Niggli reduction, the lattice vector lengths satisfy $a \le b \le c$. Thus, assuming a uniform relative error across lattice parameters, $c$ will exhibit the largest absolute error, as it corresponds to the longest Bravais lattice vector in the Niggli-reduced cell. However, assuming a uniform relative error across lattice parameters is a strong simplification, and as can be seen from the analysis of MAE by crystal system, there is likely significant interaction between model architecture and crystal system during the task of reconstruction. 

Moreover, for each model, the lattice-angle KLD tends to be higher on the Alexandria DS-A/B dataset than on the JARVIS Supercon-3D benchmarks. Two hypotheses emerge for explaining this effect. First, models may generalize more poorly as training and test set data point counts increase, but this is likely not the case, as it conflicts with empirically established neural scaling laws~\cite{Kaplan2020Scaling}. Second, and more plausibly, the increased diversity of represented crystal systems in the JARVIS Supercon-3D dataset (Figure~\ref{fig:dataset-statistics}) may allow models to generalize better to unknown lattice angles. This is yet to be investigated, but if correct, this could become a useful principle for increasing reconstruction fidelity in general.

For all models and datasets, we compare the collective performance at reconstructing crystals of differing crystal systems. When all {twelve} benchmarks are pooled by crystal system (Figures~\ref{fig:crystal_system_mae_lengths} and \ref{fig:crystal_system_mae_angles}; Table~\ref{tab:crystal_system_summary}), the reconstruction error is clearly non-uniform across crystal systems. Cubic systems exhibit the lowest pooled lattice-length MAE, while tetragonal systems also perform relatively well and show the lowest pooled angular MAE. Hexagonal, trigonal, and orthorhombic systems tend to display larger lattice-length errors, particularly in the 
$c$ parameter, and monoclinic systems show the largest angular errors overall. These trends are broadly consistent with the idea that crystal systems with stronger geometric constraints and fewer independent lattice degrees of freedom are easier to reconstruct, although symmetry alone does not fully explain the results. Because these statistics are pooled across models and datasets, and because some crystal systems are represented by relatively few samples (especially monoclinic), they should be interpreted as aggregate trends rather than definitive estimates of the intrinsic difficulty of a given crystal system.

There also appears to be a tradeoff between reconstruction fidelity and computational cost. Although computational cost is not the primary focus of this paper, it is one of the primary motivations for the widespread adoption of machine learning in computational materials discovery. Thus, we report total parameter count, total epochs, time per epoch, and time per inference for each model benchmarked in this study in Table~\ref{tab:computationalcost}. {At inference, FlowMM was the fastest model, followed by CDVAE, AtomGPT, and MatterGen, whereas MatterGen had the shortest training time per epoch.} Under the assumption that the inference conditional entropy does not have a strong relationship with reconstruction fidelity (see the Discussion section), these results suggest that each model may be preferable depending on the target application. {For extremely quick sampling of the crystal search space, FlowMM is the most favorable owing to its low inference cost, albeit at the expense of reconstruction fidelity. If one desires high-fidelity lattice-parameter prediction at a moderate computational cost, CDVAE is preferable owing to its low relative lattice error. Finally, if one prioritizes atomic-coordinate fidelity, MatterGen is most favorable, achieving the lowest coordinate error of any benchmarked model, with AtomGPT a close second; however, MatterGen incurs the highest per-inference cost, and AtomGPT's accuracy comes at the cost of substantially lower throughput and a far larger parameter count than FlowMM and CDVAE.}

Regarding acceptable error scales, sub-angstrom lattice MAE with single-digit-degree angle MAE is generally consistent with seed-quality unit cells, and a mean Cartesian RMSD below 0.1 Å already indicates high-fidelity atomic placement. Once angular errors drift into the high teens and lattice-length MAE approaches about 1 Å, the cell is usually distorted enough that the result should be treated as a coarse proposal rather than a faithful reconstruction. Therefore, in this benchmark, {CDVAE, AtomGPT, and MatterGen} models are likely accurate enough for practical screening or initialization of candidate structures for DFT, with CDVAE preferred for lattice accuracy and {MatterGen (closely followed by AtomGPT)} preferred for coordinate accuracy, while FlowMM is mainly justified when throughput matters more than cell fidelity.

\section{Discussion}

As shown in Table~\ref{tab:architecture}, the four models differ in the amount of prior information they receive during reconstruction, and these input differences may relate to the performance gap observed in the results. Thus, we emphasize that our comparison does not attempt to rank architectures universally but rather illustrates how reconstruction fidelity varies under different inference information regimes.

From an information-theoretic view, every crystal structure is specified by a finite amount of information, and these generative models are high-dimensional conditional probability distributions from which crystal reconstructions are sampled given some quantity of information about the target crystals. Since the models are supplied with different amounts of information before reconstruction, the total information they recover is inversely related to the amount they start with.  {In information theory, the conditional entropy of a random variable quantifies how much uncertainty remains about one variable once another is known \cite{cover2006elements}. By default, the CDVAE crystal embedding vector has a dimension of 256, and we use this value in the AtomBench benchmarks (the CDVAE hyperparameter table is provided in the Appendix); because this latent representation is intended to encode structural information from the target crystal, we posit that a 256-dimensional embedding may be sufficient to retain substantial stoichiometric, coordinate, and lattice information, which in turn may give CDVAE an advantage in crystal reconstruction relative to the other models and may lead to lower inference uncertainty than FlowMM for a given target crystal, all else being equal. {The remaining models do not receive such an embedding and instead condition only on composition, optionally augmented by T$_c$. As summarized in Table~\ref{tab:regimes}, this organizes the benchmarks into three conditioning-information regimes, and because the models within a given regime are supplied with the same information about the target crystal, we treat their conditional entropies as approximately equal within that regime: a composition-only regime (FlowMM, AtomGPT, and MatterGen, conditioned on $\mathbf{A}$), a composition-plus-T$_c$ regime (AtomGPT Tc and MatterGen Tc, conditioned on $\mathbf{A}$ and T$_c$), and the embedding regime occupied by CDVAE. Under the assumption that conditioning on T$_c$ is informative and therefore lowers the conditional entropy, one would expect the inference conditional entropies $H$ of each model to obey}}
\begin{equation}
\begin{aligned}
\text{stoichiometry: }\quad
  & H_{FlowMM}(\mathbf{M}\mid\mathbf{A})
    \approx H_{AtomGPT}(\mathbf{M}\mid\mathbf{A})
    \approx H_{MatterGen}(\mathbf{M}\mid\mathbf{A}), \\
\text{stoichiometry}+T_c\text{:}\quad
  & H_{AtomGPT}(\mathbf{M}\mid\mathbf{A},T_c)
    \approx H_{MatterGen}(\mathbf{M}\mid\mathbf{A},T_c), \\
\text{crystal embedding:}\quad
  & H_{CDVAE}(\mathbf{M}\mid f^{\theta}(\mathbf{A},\mathbf{X},\mathbf{L}))
    \lesssim H(\mathbf{M}\mid\mathbf{A}), \\
\text{if $T_c$ informative:}\quad
  & H(\mathbf{M}\mid\mathbf{A},T_c)
    \lesssim H(\mathbf{M}\mid\mathbf{A}).
\end{aligned}
\end{equation}

where $\mathbf{A}$ is the set of atomic element labels, $\mathbf{X}$ is the set of atomic coordinates, $\mathbf{L}$ is the set of lattice parameters, $f^\theta$ is the CDVAE encoder acting to produce a latent embedding vector $\mathbf{z}$, and $\mathbf{M}$ is the target crystal structure.  {We use conditional entropy as a conceptual proxy for the amount of unresolved structural information at inference time, rather than as a directly estimated quantity. {Because the models within the composition-only regime are conditioned on identical information, these inequalities make FlowMM, AtomGPT, and MatterGen directly comparable to one another, with CDVAE set apart as the lower-entropy embedding case. Comparisons that cross regimes, for example CDVAE against the composition-plus-T$_c$ models, remain ambiguous, since it is not immediately obvious whether T$_c$ conditioning or an embedding of the entire target crystal provides more information. Moreover, it follows that lower inference conditional entropies, i.e., lower prediction uncertainty due to known information, could bias models to exhibit greater crystal reconstruction fidelity. {Across regimes, reconstruction fidelity is broadly consistent with this picture: for lattice-parameter prediction, the embedding model CDVAE most strongly outperforms FlowMM, with AtomGPT and MatterGen intermediate. Within the composition-only regime, however, FlowMM, AtomGPT, and MatterGen are given identical information yet differ sharply in fidelity (for atomic-coordinate prediction, MatterGen and AtomGPT both strongly outperform FlowMM as well as CDVAE), so these within-regime gaps cannot be explained by information content and must instead reflect differences in architecture and pretraining.} The magnitude of the influence of the information discrepancy is currently undetermined, and it is possible that the influence of the information discrepancy may have differing magnitudes for lattice parameter and atomic coordinate prediction. {We emphasize that we do not claim inference conditional entropy to be a more influential determinant of reconstruction fidelity than architectural
  choice; the present data do not resolve their relative contributions, and quantifying them is left to future work.}
}

{Two comparisons within these matched regimes are especially informative. The first is MatterGen versus AtomGPT. Both models were pretrained and, in both the stoichiometry-only and stoichiometry and T$_c$ regimes, condition on identical information, so the comparison between them isolates the effect of MatterGen's domain-specific pretraining on crystal structures, in contrast to AtomGPT's general-purpose, text-pretrained base model, together with its physically informed, equivariant inductive bias in the architecture of MatterGen. Despite having roughly $140\times$ fewer parameters than AtomGPT (Table~\ref{tab:computationalcost}), more than two orders of magnitude, MatterGen matches or exceeds AtomGPT on atomic-coordinate fidelity (Table~\ref{tab:reconstruction_all}) and on match rate. That a model this much smaller can equal or surpass a generalist of AtomGPT's scale is itself strong evidence that the combination of domain-specific pretraining and physical inductive bias has a powerful positive effect on reconstruction fidelity, effectively substituting for orders of magnitude more parameters. The second comparison concerns FlowMM, which shares the composition-only regime but, unlike AtomGPT and MatterGen, is not pretrained on a large dataset. Its substantially weaker reconstruction, therefore, cannot be cleanly attributed: the deficit is confounded by its flow-matching architecture and the absence of pretraining, and the present data do not allow us to isolate these two factors. We accordingly refrain from concluding that FlowMM's architecture is inherently limiting; disentangling architecture from pretraining is left to future work.}

 {
{
As a direct probe of this picture, we supplied T$_c$ explicitly to the two models that support it (AtomGPT and MatterGen) and compared each against its unconditioned counterpart. Reconstruction fidelity did not consistently improve, indicating that the performance gaps we observe are driven more by the structural information available at inference (e.g., the latent crystal embedding seen by CDVAE) than by a scalar property label alone. In the language of the regimes above, this means the hypothesized inequality $H(\mathbf{M}\mid\mathbf{A},T_c)\lesssim H(\mathbf{M}\mid\mathbf{A})$ is not borne out empirically: moving a model from the composition-only regime to the composition-plus-T$_c$ regime changes its conditioning information but not its reconstruction error in a consistent direction. For example, MatterGen's coordinate ccRMSD rises slightly from $0.238$ to $0.244$ on Alexandria and from $0.485$ to $0.499$ on JARVIS, whereas AtomGPT's falls from $0.826$ to $0.767$ on JARVIS. A scalar property label therefore cannot be assumed to be an informative conditioning variable for crystal reconstruction before the relevant data has been collected; whether T$_c$ defines a genuinely lower-entropy regime is an empirical question rather than an assumption.}
}

{Taken together, these analyses illustrate why conditional-entropy regimes are a useful organizing principle for reconstruction benchmarking. Grouping models by their conditioning information lets us hold that information fixed and ask what remains: here, that T$_c$ does not reliably reduce reconstruction uncertainty, that domain-specific pretraining and physical inductive bias can outweigh a hundredfold parameter advantage, and that FlowMM's architecture and lack of pretraining remain entangled and cannot yet be separated. Without this regime structure, such effects would be confounded with the unequal information each model receives, and differences in reconstruction fidelity could be misattributed to architecture alone.}

\subsection{Future Work}

Future work should supply each model with equivalent information prior to reconstruction, enabling fine-grained architectural comparisons without the confound of unequal inference conditional entropy when reconstructing crystals. AtomGPT and MatterGen already support property conditioning, whereas FlowMM and CDVAE would require architectural modifications to condition on a scalar property such as T$_{\mathrm c}$; we sketch concrete constructions for both in the Appendix. 

{A second orthogonal modification targets the pretraining confound identified in the Discussion.} Because FlowMM is the only benchmarked model that is not generatively pretrained, its weaker reconstruction within the composition-only regime cannot be cleanly attributed to its architecture, as architecture and the absence of pretraining are entangled. A natural way to disentangle them is to endow FlowMM with a generative pretraining stage analogous to those of AtomGPT and MatterGen, for instance pretraining the flow on a large corpus of stable crystal structures before fine-tuning on the superconductor datasets, and then re-benchmarking it within the same composition-only regime. If a pretrained FlowMM closed much of the present gap, the deficit would be attributable largely to pretraining rather than to flow matching as an architecture; if a substantial gap remained, the evidence would instead point to the architecture itself. Either outcome would sharpen the architecture-versus-pretraining comparison enabled by the matched-regime analysis.}

Moreover, for these models to have the broadest impact on superconductor discovery, the superconducting temperatures predicted for crystals by these models must be physically valid. An orthogonal model benchmarking framework would evaluate the T$_c$ of the predicted crystals using first-principles methods or surrogate forward models; however, this is outside the scope of this paper, and we designate it as future work.

{A broader question concerns whether AtomBench generalizes beyond superconductors, and whether distribution shifts observed in superconductor datasets would be recapitulated for other crystal properties. We note that resolving this by studying a single property is fundamentally insufficient: any observed distribution shift in a superconductor benchmark cannot be unambiguously attributed to the superconducting property per se, because superconducting datasets are systematically co-distributed with correlated latent properties such as Debye temperature, stoichiometry, and crystallographic symmetry trends. Disentangling property-specific effects from these confounds requires extending the benchmark across multiple property domains. To this end, AtomBench is designed to be property-agnostic, and the present superconductor study is the first in a planned series of benchmarks that span other crystal properties available in JARVIS, including the dielectric constant and Debye temperature. {Because the present datasets comprise BCS (electron-phonon-mediated, metallic) superconductors, we expect the reconstruction behavior reported here to translate reasonably to other metallic systems, though it may differ for non-metallic systems.} The superconducting transition temperature was selected as the initial conditioning variable for two reasons. First, the discovery of superconductors is among the highest-priority applications of inverse design in materials science. Second, $T_c$ is a scalar, making it the simplest possible conditioning variable for a generative model such as AtomGPT; higher-order properties, such as XRD spectra or magnetic anisotropy tensors, introduce additional architectural complexity that is more tractable to address once the benchmarking framework has been validated on simpler targets.}

\section{Conclusion}
\label{sec:conclusion}

In this work, we introduce AtomBench, an extendable, task-specific generative crystal structure model benchmarking framework for the task of superconducting crystal structure reconstruction. We benchmark AtomGPT, CDVAE, FlowMM, and MatterGen, four generative models representing three model families, on two superconductivity datasets, JARVIS Supercon-3D and Alexandria DS-A/B. In this reconstruction setting, CDVAE achieves the best lattice-parameter fidelity, whereas MatterGen attains the lowest atomic-coordinate error, closely followed by AtomGPT; FlowMM shows the largest lattice and lattice-distribution errors under our evaluation metrics. {We further introduce the continuous corrected RMSD (ccRMSD), a continuous, match-rate-independent measure of atomic-coordinate fidelity that is defined for every reconstructed structure rather than only those passing a hard structure-matching tolerance; evaluated this way, the four models separate along a continuum rather than into two discrete regimes, indicating that the apparent clustering produced by the conventional matched RMSD is partly an artifact of the matching cutoff.} We hypothesize that a major confound in current reconstruction benchmarks is the mismatch in conditioning information provided at inference time. Models that are given more informative conditioning variables (e.g., target property labels or latent embeddings derived from the target structure) may face lower effective reconstruction uncertainty than models conditioned only on atomic species. Consequently, observed differences in reconstruction fidelity may not be attributable solely to architectural differences. Future work will focus on comparing model architectures under experimental conditions in which each receives an equivalent amount of information about the target crystal. Establishing such parity will enable a rigorous evaluation of the intrinsic inductive biases that underlie each model’s design. {Toward that end, the conditioning-information regimes introduced here already let us separate some effects from others: holding conditioning information fixed within the composition-only regime, MatterGen matches or exceeds AtomGPT on atomic-coordinate fidelity with more than two orders of magnitude fewer parameters, evidence that domain-specific pretraining and physically informed inductive bias can substitute for sheer model scale, whereas FlowMM's architecture-versus-pretraining confound remains a target for future work.} {We release the analysis framework as an open-source, model-agnostic Python package, \texttt{atombench}, that reproduces the reconstruction-accuracy metrics, figures, and tables reported here from a single benchmark file and lets users submit their own results to the JARVIS-Leaderboard; although we demonstrate it on four models, any inverse model that produces crystal reconstructions can be substituted.} All configurations, scripts, and datasets used for this study are openly accessible and can be found in the subsequent data availability section.

\section{Data Availability}
The datasets used in this work are available at \url{https://doi.org/10.6084/m9.figshare.6815699} and \url{https://doi.org/10.6084/m9.figshare.31045597}. The code used in this study is publicly available at {\url{https://github.com/atomgptlab/atombench}, and the analysis package can be installed from PyPI via \texttt{pip install atombench}}.

\section{Author Contributions}
K.C. conceived the project. C.C. and K.C. developed the software pipeline and performed the computational experiments. C.C. carried out data curation, validation, and visualization. A.H.R., K.C., and C.C. performed the formal analysis and interpreted the results. C.C. and A.H.R. wrote the original draft of the manuscript. All authors reviewed and edited the manuscript. A.H.R. and K.C. supervised the project and secured funding.

\section*{Conflict of Interest}
The authors declare no conflict of interest.

\section*{Acknowledgements}
We thank the AtomGPTLab cluster for computational support.
We thank the Pittsburgh Supercomputer Center (Bridges2) and the San Diego Supercomputer Center (Expanse) through allocation DMR140031 from the Advanced Cyberinfrastructure Coordination Ecosystem: Services \& Support (ACCESS) program, which is supported by National Science Foundation grants \#2138259, \#2138286, \#2138307, \#2137603, and \#2138296. 
We also recognize the computational resources provided by the WVU Research Computing Dolly Sods HPC cluster, which is funded in part by NSF OAC-2117575.
Support for the West Virginia group came from the West Virginia Higher Education Policy Commission under the call Research Challenge Grant Program 2022, Award RCG 23-007.

\bibliography{crystalgen_references}

\clearpage
\appendix

\section{Appendix}
\label{appendix}

\subsection{Density Functional Theory and Inverse Design of Superconductors}
\label{app:background}
Electrons in crystalline solids can organize into remarkable collective states. High-temperature superconductivity is a prime example that lies at the forefront of condensed-matter physics and materials engineering. 
The conventional \textit{in-silico} discovery pipeline relies heavily on density functional theory (DFT). DFT computes structure-property relationships from first principles by solving approximations to the many-body Schrödinger equation, enabling the prediction of observables such as formation energy, band gap, elastic moduli, and superconducting critical temperature\cite{gross2013density,choudhary2021atomistic}. DFT and its extensions, such as superconducting DFT (SCDFT), are reliable for computing superconducting critical temperatures when superconductivity is driven by electron-phonon coupling, as in conventional superconductors~\cite{giustino2017electron, oliveira1988density, luders2005ab}. However, these computational methods become problematic for unconventional superconductors such as cuprates and iron-based compounds, where strong electron-electron correlations play a major role, and no consensus exists on a fully predictive ab initio theory~\cite{furness2018accurate, pokharel2022sensitivity, kent2008combined}. Moreover, standard DFT methods can underestimate critical temperatures by significant margins. For example, nearly 50\% in some hydride systems, highlighting the need for correction schemes or more advanced formalisms~\cite{chen2025impact, held2008bandstructure, selisko2024dynamical}. Though accurate in many cases, DFT typically requires substantial computational effort per structure, scaling cubically with system size, thereby limiting throughput to only a few candidates at a time. Despite the computational cost, DFT-based methods remain the most widely used and trusted approach for describing conventional (electron-phonon-mediated) superconductors, where the pairing mechanism is well understood and reliably captured within current first-principles frameworks.

While theoretical efforts focus on developing more accurate or more explicit theoretical models for high T$_c$ superconductivity, the use of artificial intelligence methods can be a game-changer, as they provide extrapolation techniques capable of deriving hitherto unknown information from the dataset~\cite{choudhary2022recent}. By training on well-characterized DFT corpora with specified pseudopotentials, convergence settings, and other specified methodologies, generative models inherit the DFT configuration’s physical constraints, chemical validity assumptions, and systematic biases while sharply lowering the cost of candidate generation and property prediction. As a surrogate of the DFT-induced distribution, the model cannot, without further correction or transfer learning, achieve accuracy beyond that of its teacher. 
The goal is not to replace DFT but to use machine learning to propose high-quality candidates that are likely to exhibit the desired properties. These candidates can then be validated using more precise, although computationally expensive, \textit{ab initio} methods.
Moreover, machine learning can be used to quickly estimate properties at high throughput when it would be too costly to do so using DFT.

With that said, high-quality DFT databases form the structural backbone for machine-learning-driven studies of conventional (electron-phonon-mediated) superconductors. For example, the JARVIS‑DFT infrastructure contains over 90,000 materials with extensive computed properties, ranging from structural, electronic, and mechanical to phonon-related data, including a refined subset of about 1,058 superconductors characterized via electron-phonon coupling and the 
McMillan-Allen-Dynes formula to estimate T$_c$~\cite{choudhary2020joint, choudhary2025jarvis, choudhary2022designing,choudhary2024jarvis, wines2023recent}. The electronic-structure data in JARVIS-DFT have also been used to train fast models, such as SlaKoNet, a neural Slater-Koster tight-binding framework that predicts band structures and band gaps across the periodic table~\cite{Choudhary2025SlaKoNet}. Similarly, high-throughput screening via machine learning and DFT within JARVIS-DFT has evaluated over 1,000 two-dimensional materials, yielding 34 dynamically stable superconductors with T$_c > 5\ \mathrm{K}$~\cite{wines2023high} and high-pressure hydride superconductors \cite{wines2024data}.
Beyond this, the Alexandria database, curated by Marques and collaborators, provides an even broader foundation, now including more than 4.4 million inorganic compounds across multiple dimensionalities (3D, 2D, 1D)~\cite{schmidt2024improving}, with computed properties accessible under a permissive open license. This vast repository enables machine-learning-accelerated workflows that have already suggested promising hydride superconductors among more than one million candidate compounds~\cite{sanna2024prediction, gao2025enhanced, cerqueira2024sampling}.

Machine learning for high-throughput materials screening can be broadly separated into two categories: forward design and inverse design. Forward design, also known as the direct or predictive problem, involves determining a material’s macroscopic properties based on a complete specification of its atomic crystallographic structure. In the context of crystalline materials, essential information is provided by the Bravais lattice vectors and the atomic positions within the unit cell. Alternatively, the system can be defined by the Wyckoff positions, cell parameters, and the space group. In both cases, it is also necessary to specify the chemical identities of the constituent elements. Approaches such as graph neural networks (GNNs) and equivariant message-passing networks are trained on existing DFT databases to learn the complex, nonlinear relationship between crystal structure/property pairs. These forward models encode the atomic species and spatial arrangement into a structured representation (often referred to as a crystal graph) and approximate the function $
f: \bigl(\mathbf{A},\mathbf{X},\mathbf{L}\bigr) \;\mapsto\; \mathbf{y}
$,
where $\mathbf{A}$ represents the atomic species, $\mathbf{X}$ the atomic coordinates, $\mathbf{L}$ the set of lattice vectors, and $\mathbf{y}$ the target property vector. Once trained, such models can evaluate millions of hypothetical structures in minutes to hours, making them suitable for high-throughput virtual screening and accelerating the discovery pipeline by several orders of magnitude.

Conversely, inverse design poses a complementary and substantially more challenging problem: starting from a desired property vector $\mathbf{y}$, identify one or more crystal structures $\mathbf{M}=(\mathbf{A},\mathbf{X},\mathbf{L})$ that are likely to realize it. This inverse problem is inherently ill-posed; the structure-property mapping is many-to-one, nonlinear, and discontinuous, meaning that no unique or closed-form inverse exists. Recent advances in generative modeling offer a probabilistic path forward by learning the conditional distribution
$ p_{\boldsymbol{\phi}}\!\bigl(\mathbf{A},\mathbf{X},\mathbf{L}\mid\mathbf{y}\bigr)
$,
which can be sampled to generate chemically valid, symmetry-consistent crystals that are biased toward the target properties. Similar to forward models, these inverse models are trained on large corpora of crystal structure/property pairs generated from experimental or DFT methods, optimizing likelihood or divergence-based objectives to align the learned generative distribution with the empirical one. Once trained, inverse design reduces to sampling from 
the learned conditional distribution $p_{\boldsymbol{\phi}}$, conditioned on the vector of properties $\mathbf{y}^{\ast}$, to generate crystal structures.

\subsection{AtomGPT}
\label{atomgpt}

AtomGPT is a generative, pretrained transformer (GPT) adapted to predict crystal structures and target properties via text generation. This framing not only enables the model to predict lattice parameters and atomic coordinates by generating text sequences but also makes the interface more intuitive for researchers, since structures can be queried and generated in natural language rather than specialized code. Since this study focuses on testing AtomGPT's inverse design accuracy, its ability to predict target properties as a function of lattice structure will not be discussed; we will instead discuss its ability to predict lattice structures as a function of a desired property.

At its core, AtomGPT utilizes a pretrained language model; for this study, the Mistral-7b-BNB-4bit GPT was used for its low computational cost and strong performance on benchmarks~\cite{jiang2023mistral7b}. Out of the box, this pretrained language model is a generalist and has no particular expertise in crystal and materials problems compared with other problem domains. However, there is strong evidence that resuming model training (finetuning) on crystal structures strongly improves the performance of language models for predicting material properties and lattice structures~\cite{gruver2025finetunedlanguagemodelsgenerate,Choudhary2023AtomGPT}. To perform fine-tuning, AtomGPT uses the Hugging Face SFTTrainer to process a dataset of crystal-property pairs and update the model weights accordingly. The crystal-property pairs used for fine-tuning follow a common textual schema, and the following example illustrates how they are represented as text during training.

\vspace{10mm}
\begin{center}
\begin{minipage}{0.92\linewidth}
\begin{verbatim}
Input:  The crystal's chemical formula is Nb3Sn, and the superconducting
        transition temperature is 18.3 K. Generate atomic structure 
        description with lattice lengths, angles, coordinates and atom types.

Output:    5.32 5.32 5.32
            90 90 90
            Sn 0.000 0.000 0.000
            Nb 0.000 0.500 0.500
            Nb 0.500 0.000 0.500
            Nb 0.500 0.500 0.000
\end{verbatim}
\end{minipage}
\end{center}
\vspace{10mm}

During finetuning, the model generates a predicted crystal structure for each input prompt. The cross-entropy loss is then computed token-by-token against the textual encoding of the corresponding ground-truth structure, and model weights are updated via gradient descent to minimize this loss. After finetuning, inference is performed using the finetuned checkpoint loaded via the Hugging Face Transformers library.
\begin{table}[H]
\centering
\label{tab:agpt_hparams}
\begin{tabular}{p{0.34\linewidth} p{0.60\linewidth}}
\hline
\textbf{Parameter} & \textbf{Value} \\
\hline
Model & \texttt{unsloth/mistral-7b-bnb-4bit} \\
Epochs & 10 \\
Batch size (global) & 2 \\
Per-device train batch size & 16 \\
Gradient accumulation steps & 1 \\
Learning rate & \(2\times 10^{-4}\) \\
Optimizer & AdamW (8-bit) \\
LR scheduler & Linear \\
Max sequence length & 2048 \\
Test ratio & 0.1 \\
Seed & 3407 \\
Load in 4-bit & \texttt{true} \\
Quantization mode & \texttt{bnb-4bit} \\
Alpaca-style prompt & \texttt{Instruction:\{\}  Input:\{\}  Output:\{\}} \\
Instruction prompt & Below is a description of a superconductor material. \\
Output prompt & Generate a correct corresponding bulk crystal POSCAR with lattice lengths, angles, coordinates and atom types. \\
\hline
\end{tabular}
\captionsetup{justification=raggedright,singlelinecheck=false}
\caption{AtomGPT hyperparameters and implementation details.}
\end{table}

The authors of this paper are the developers of AtomGPT, so no repository fork was used to compute the benchmarks. The repository can be found at \url{github.com/atomgptlab/atomgpt}. 
\\
\\
AtomGPT Tutorial Notebook:
\url{github.com/knc6/jarvis-tools-notebooks/blob/master/jarvis-tools-notebooks/AtomGPT_example.ipynb}

\subsection{CDVAE}
\label{cdvae}

CDVAE is an inverse model for materials design that generates periodic and physically plausible materials using a diffusion variational autoencoder architecture. It has three main functionalities: reconstruction, generation, and property optimization. For this study, only the reconstruction task was used; therefore, we will limit our discussion of CDVAE to reconstruction and concepts relevant to it. Under the hood, CDVAE reconstructs materials using a three-step process. First, an SE(3)-equivariant periodic graph neural network encoder $PGNN_{ENC} (\mathbf{M})$ maps a crystal structure $\mathbf{M}$ to a latent representation $\mathbf{z}$. Importantly, the crystal structure $\mathbf{M}=(\mathbf{A},\mathbf{X},\mathbf{L})$ is fully described by three lists, $\mathbf{A}$, $\mathbf{X}$, and $\mathbf{L}$, which contain the crystal’s atom types, atomic coordinates, and Bravais lattice vectors, respectively. Second, three distinct multilayer perceptrons (MLPs) map $\mathbf{z}$ to a set of three aggregated properties: the atomic composition $\mathbf{A}$, the Bravais lattice vectors $\mathbf{L}$, and the number of atoms $N$. After these aggregated properties are predicted, a provisional structure is instantiated by assigning atom types according to the predicted composition $\mathbf{A}$, placing them at uniformly sampled atomic coordinates within the unit cell defined by $\mathbf{L}$, and perturbing these positions with Gaussian noise to obtain the noisy material $\mathbf{\tilde{M}}$. Third, a conditional score-matching decoder $PGNN_{DEC}(\mathbf{M}\mid\mathbf{z})$ parameterized by an SE(3)-equivariant periodic graph neural network denoises $\mathbf{\tilde{M}}$ via annealed Langevin dynamics to produce a reconstruction of the original crystal $\mathbf{M}$. During training, the encoder, decoder, and aggregate property heads are jointly optimized with a master loss function that is a linear combination of the individual loss functions from each neural network used in the CDVAE architecture: $\mathcal{L}=\mathcal{L}_{\mathrm{AGG}}+\mathcal{L}_{\mathrm{DEC}}+\mathcal{L}_{\mathrm{KL}}$. In this expression, $\mathcal{L}_{\mathrm{AGG}}$ represents the atom type classification and lattice regression loss, $\mathcal{L}_{\mathrm{DEC}}$ represents the denoising score-matching loss, and $\mathcal{L}_{\mathrm{KL}}$ represents the Kullback-Leibler divergence regularization loss for the encoder, which is characteristic of many VAE architectures.
\begin{table}[H]
\captionsetup{justification=raggedright,singlelinecheck=false}
\caption{CDVAE hyperparameters and implementation details.}
\label{tab:cdvae_hparams}
\begin{tabular}{p{0.36\linewidth} p{0.58\linewidth}}
\hline
\textbf{Parameter} & \textbf{Value} \\
\hline
Max atoms per structure & 20 \\
Training epochs (max) & 100 \\
Early stopping patience & 5 \\
Teacher forcing (max epoch) & 20 \\
Data split (train / val / test) & 0.8 \; / \; 0.1 \; / \; 0.1 \\
Batch size (train / val / test) & 64 \; / \; 64 \; / \; 64 \\
Encoder settings & Default \\
Decoder settings & Default \\
Optimizer settings & Default \\
Training settings & Default \\
Latent Dimension & 256 \\
\hline
\end{tabular}
\end{table}

A fork of CDVAE was utilized for this study. The changes made to the original CDVAE repository include removing Weights \& Biases logging, adding configuration files for the JARVIS Supercon-3D and Alexandria DS-A/B datasets, and fixing a bug that caused erroneous values for the project root path. The fork can be found at \url{github.com/crhysc/cdvae}.
\\
\\
CDVAE Tutorial Notebook:
\url{github.com/crhysc/jarvis-tools-notebooks/blob/master/jarvis-tools-notebooks/cdvae_example.ipynb}

\newpage
\subsection{FlowMM}
\label{flowmm}

FlowMM is an inverse model for materials design that uses a Riemannian flow-matching architecture to predict stable crystal structures with known or novel compositions. FlowMM is built to perform two tasks: crystal structure prediction (CSP) and \textit{de novo} generation (DNG). For this study, only the CSP task was used, so we limit our discussion of FlowMM to CSP and concepts relevant to it. CSP is the process by which FlowMM predicts the atomic coordinates and Bravais lattice vectors of a crystal with only its composition specified, and FlowMM represents crystals by their position on a product manifold $\mathcal{C} := \mathcal{A} \times \mathcal{F} \times \mathcal{L}$. The submanifold $\mathcal{F}$ is a collection of $n \times 3$ flat tori representing the periodic space of atomic coordinates, the submanifold $\mathcal{L}$ is the space of lattice parameters $\{a,b,c\}\in \mathbb{R}^{+3}$ and $\{\alpha,\beta,\gamma\}\in[60,120]^3$, subject to the Niggli reduction \cite{niggli}. Because of domain boundaries in $\{\alpha,\beta,\gamma\}$, FlowMM represents lattice parameters in an unconstrained coordinate system via an invertible transformation $\phi$; training and inference take place in this flat space, and $\{\alpha,\beta,\gamma\}$ are recovered by applying the inverse transformation $\phi^{-1}$. The submanifold $\mathcal{A}$ represents compositions, which in CSP are fixed $h$-dimensional one-hot vectors. Since composition is specified during both training and inference, the learned vector field has no active components along $\mathcal{A}$, effectively reducing the manifold to $\mathcal{C}' := \{\mathbf{A}\} \times \mathcal{F} \times \mathcal{L}$, where $\mathbf{A} \in \mathcal{A}$. Here, the brackets denote holding a set of atomic coordinates constant while allowing the other parameters to vary freely. Because $\mathcal{F}$ and $\mathcal{L}$ are flat, closed-form geodesics exist and take the form of straight line segments; however, geodesics in the $\mathcal{F}$ submanifold wrap around due to the toroidal nature of $\mathcal{F}$.

In this space, a time-dependent flow $\psi_t$ is defined as the solution of the differential equation
\[
\frac{d}{dt}\psi_t(x) = u_t(\psi_t(x)), \quad \psi_0(x) = x,
\]
which pushes an initial density $p_{0}$ along a probability path $p_t$ to a target distribution $p_{1} = q$. The initial probability density on the manifold is comprised of the uniform distribution on $\mathcal{F}$, the LogNormal distribution for $\{a,b,c\}$, and the uniform distribution for $\{\alpha,\beta,\gamma\}$ pushed through $\phi$. In $\mathcal{F}$, the conditional targets use toroidal logarithmic displacements with the mean tangent translation across atoms removed, yielding a translation-invariant marginal path \cite{Miller2024FlowMM}. The target distribution $q$ is the empirical measure supported on the training set,
$q = \frac{1}{N}\sum_{i=1}^N \delta_{x_i},$
from which we sample $x_1$ during training.
 In practice, training proceeds by sampling an initial point $x_0 \sim p_0$, a time $t \sim$ uniformly $(0,1)$, and a training example $x_1 \sim q$. The conditional flow matching construction guarantees the existence of a velocity field $u_t(x|x_1)$ that connects $x_0$ to $x_1$ along a straight path (because $\mathcal{C}'$ is Euclidean)\cite{rfm}. A permutation-equivariant, translation-aware GNN then predicts a velocity vector $v_t^{\theta}(x)$ in the tangent space at the interpolated location $x_t$. The loss is the squared Riemannian distance between $v_t^{\theta}(x_t)$ and the target velocity $u_t(x_t|x_1)$, averaged over samples. Minimizing this objective aligns the learned velocity field with the analytic conditional vector fields, ensuring that the trained model transports the base distribution $p_0$ toward the empirical data distribution $q$. Permutation invariance follows from relabel-equivariant message passing, and rotation invariance follows from using Niggli-reduced lattice parameters in $\mathcal{L}$, so the induced density is $S_n$- and $SO(3)$-invariant by construction \cite{Miller2024FlowMM}.

At inference time, CSP reduces to integrating the learned flow forward in time: given a composition $\mathbf{A}$, a point is sampled from the base distribution $p_0$ on $\mathcal{F} \times \mathcal{L}$ and transported to $t=1$ under the learned flow. The result is a predicted set of lattice parameters and atomic coordinates consistent with the specified composition.
\begin{table}[H]
\centering
\captionsetup{justification=raggedright,singlelinecheck=false}
\caption{FlowMM hyperparameters and implementation details.}
\label{tab:flowmm_hparams}
\begin{tabular}{p{0.36\linewidth} p{0.58\linewidth}}
\hline
\textbf{Parameter} & \textbf{Value} \\
\hline
Max atoms per structure & 24 \\
Training epochs (max) & 100 \\
Early stopping patience & 5 \\
Teacher forcing (max epoch) & 20 \\
\texttt{dim\_coords} & 3 \\
Data split (train / val / test) & 0.8 \; / \; 0.1 \; / \; 0.1 \\
Batch size (train / val / test) & 64 \; / \; 64 \; / \; 64 \\
Vector field network & Default \\
Model settings & Default \\
Optimizer settings & Default \\
\hline
\end{tabular}
\end{table}

A FlowMM fork was used in this study. The changes made to the original FlowMM repository include removing Weights \& Biases logging, adding configuration files for the JARVIS Supercon-3D and Alexandria DS-A/B datasets, and modifying the FlowMM hard code to accept these datasets. FlowMM was not shipped to automatically enable users to train on arbitrary datasets, but the code modifications discussed in our FlowMM tutorial notebook explain the changes necessary to compute these benchmarks.
\\
\\
FlowMM Tutorial Notebook:
\url{github.com/crhysc/jarvis-tools-notebooks/blob/master/jarvis-tools-notebooks/flowmm_example.ipynb}

\newpage
\subsection{MatterGen}
\label{mattergen}

MatterGen is an inverse model for materials design that generates inorganic crystal structures via a joint diffusion process tailored to the geometry of each crystallographic component~\cite{Zeni2025MatterGen}. MatterGen can perform both crystal structure prediction (CSP) and \textit{de novo} generation; for this study, only CSP was used, so we constrain our discussion accordingly. Like the other models in this study, MatterGen represents a crystal by its unit cell $\mathbf{M} = (\mathbf{A}, \mathbf{X}, \mathbf{L})$, where $\mathbf{A}$, $\mathbf{X}$, and $\mathbf{L}$ denote atom types, atomic coordinates, and Bravais lattice vectors, respectively. A separate forward diffusion process is defined for each component: coordinate diffusion employs a wrapped Normal distribution that respects the periodicity of $\mathbf{X}$ and converges to the uniform distribution at high noise; lattice diffusion takes a symmetric form and converges to a distribution centered on a cubic lattice at the average training-set atomic density; and atom-type diffusion operates in categorical space, corrupting element identities into a masked absorbing state. Given a corrupted structure at timestep $t \in \{1, \ldots, T\}$, an SE(3)-equivariant score network jointly denoises $\mathbf{A}$, $\mathbf{X}$, and $\mathbf{L}$; this constitutes the MatterGen base model, pre-trained on a large collection of stable crystal structures. For CSP, $\mathbf{A}$ is held fixed at inference, and only $\mathbf{X}$ and $\mathbf{L}$ are initialized from noise and iteratively refined over $T$ reverse diffusion steps. To steer generation toward a target superconducting transition temperature $T_c$, lightweight adapter modules inserted at each layer of the base model are fine-tuned on labeled data; at inference, classifier-free guidance uses the encoded $T_c$ to bias the denoising trajectory.
\begin{table}[H]
\centering
{
\captionsetup{justification=raggedright,singlelinecheck=false}
\caption{MatterGen hyperparameters and implementation details.}
\label{tab:mattergen_hparams}
\begin{tabular}{p{0.40\linewidth} p{0.54\linewidth}}
\hline
\textbf{Parameter} & \textbf{Value} \\
\hline
Base model & \texttt{mattergen\_base} (pretrained, adapter fine-tuned) \\
Training epochs (max) & 200 \\
Per-device train batch size & 128 \\
Gradient accumulation steps & 4 \\
Effective batch size & 512 \\
Learning rate & \(5\times 10^{-6}\) \\
Optimizer & Adam \\
LR scheduler & ReduceLROnPlateau \\
Data split (train / val / test) & 0.8 \; / \; 0.1 \; / \; 0.1 \\
Adapter / model settings & Default \\
\hline
\end{tabular}
}
\end{table}

{A fork of MatterGen was utilized for this study. The changes made to the original MatterGen repository are removing Weights \& Biases logging, adding configuration files for the JARVIS Supercon-3D and Alexandria DS-A/B datasets, and fine-tuning the pretrained \texttt{mattergen\_base} checkpoint in crystal-structure-prediction mode (composition fixed; atomic coordinates and lattice denoised). The fork can be found at \url{github.com/crhysc/mattergen}.}

\subsection{JARVIS Supercon-3D}
This was the first full database of atomic-structure information for superconductors. The authors start from 55{,}723 JARVIS-DFT structures and pre-screen using the Debye temperature \(\theta_D\) (derived from elastic tensors) and the electronic DOS at the Fermi level, retaining materials with \(\theta_D > 300\) K (5{,}618 remain) and \(N(0) > 1\) states\,eV\(^{-1}\) per valence electron (1{,}736 remain). To keep DFPT tractable, they then restrict to primitive cells with \(\le 5\) atoms, yielding 1{,}058 candidates. For these, electron-phonon coupling (EPC) is computed via DFPT in Quantum ESPRESSO with GBRV pseudopotentials \cite{giannozzi2009quantum} and the PBEsol exchange-correlation functional \cite{Perdew2008PBEsol}; T\(_c\) is estimated using the McMillan-Allen-Dynes equation with \(\mu^{*} = 0.09\), where $\mu^{*}$ is the Morel-Anderson effective Coulomb pseudopotential\cite{Morel1962}. Dynamical stability is assessed from the DFPT phonon spectra: a material is labeled “stable” only if no imaginary (negative) phonon frequencies appear at any sampled \(\mathbf{q}\)-point across the Brillouin zone (i.e., all mode eigenfrequencies satisfy \(\omega_{qj}^2 > 0\) within numerical tolerances). This yields 626 dynamically stable structures, of which 105 have T\(_c \ge 5\) K. The EPC convergence strategy is deliberately lightweight: reuse the JARVIS-converged \(k\)-point meshes, employ at least a \(2\times 2\times 2\) \(q\)-mesh, and apply Gaussian broadening \(\approx 0.05\) Ry to stabilize \(\lambda\) at modest cost. In our benchmarks, we use the full 1{,}058-structure set to evaluate property-conditioned reconstruction; if prioritizing dynamical stability, one could restrict to the 626-structure stable subset.

\subsection{Alexandria DS-A/B}
Subsequently, the Alexandria DS-A/B dataset~\cite{CerqueiraHydrideSupercons} was developed utilizing a high-throughput methodology enhanced by machine learning techniques. This comprehensive dataset comprises 8{,}253 well-converged EPC entries, including compounds such as nitrides, hydrides, and intermetallics. The source structures are screened from the Alexandria database for metallic, non-magnetic compounds on or near the convex hull (typically \(E_{\mathrm{hull}}<50\) meV/atom, with a secondary sweep allowing \(50\!\le\!E_{\mathrm{hull}}\!<\!100\) meV/atom for small cells), excluding semiconductors, insulators, semimetals, and very low density of states entries. A Debye-temperature filter T\(_D>300\) K (estimated via an ALIGNN model\cite{choudhary2021atomistic}) is applied, and structural complexity is limited to \(\le 8\) atoms per primitive cell with space-group number \(\ge 100\) (favoring tetragonal and cubic lattices). All calculations use PBEsol\cite{Perdew2008PBEsol} in Quantum ESPRESSO with PseudoDojo pseudopotentials, tight stopping criteria on energies/forces/stresses (\(10^{-8}\) a.u., \(10^{-6}\) a.u., \(5\times 10^{-2}\) kbar), a double-grid electron-phonon coupling (EPC) strategy, and Methfessel-Paxton smearing \(\approx 0.05\) Ry. Dynamic stability is assessed from DFPT phonons; a practical tolerance allows for up to three small-magnitude imaginary modes at \(\Gamma\) (\(\lesssim 35\) cm\(^{-1}\)) to account for numerical artifacts, with false positives removed at higher accuracy. Superconducting labels report the dimensionless EPC strength \(\lambda\), the logarithmic-average phonon frequency \(\omega_{\log}\), and T\(_c\) from the McMillan-Allen-Dynes formula using \(\mu^{*}=\!0.10\) (where $\mu^{*}$ is the Morel-Anderson effective Coulomb pseudopotential). In this work, we concatenate the original Alexandria DS-A (training) and Alexandria DS-B (validation) splits into a single Alexandria DS-A/B dataset for model training and evaluation. 

\subsection{Property-Conditioned Extensions of FlowMM and CDVAE}
\label{app:conditioning-extensions}

Here we expand on the Future Work proposal to give FlowMM and CDVAE (the two benchmarked models that do not natively condition on a scalar property) the ability to map a specified composition $\mathbf{A}$ and superconducting transition temperature T$_{\mathrm c}$ to atomic coordinates $\mathbf{X}$ and lattice vectors $\mathbf{L}$. For FlowMM, one potential approach involves the introduction of a submanifold characterized by a scalar property, denoted as $\mathcal{T}=\mathbb{R}^+$, where $\mathcal{T}$ represents the set of all superconducting temperatures and $\mathbb{R}
^+$ is the set of positive real numbers. Currently in FlowMM, crystals are represented on a  product manifold $\mathcal{C} = \mathcal{A} \times \mathcal{F} \times \mathcal{L}$ as described in the Appendix. To wrap $\mathcal{T}$ into the existing product manifold, one would simply use the cartesian product operation that currently links the other submanifolds to define a T$_c$-aware product manifold $\mathcal{C}' = \mathcal{A} \times \mathcal{F} \times \mathcal{L} \times \mathcal{T}$. This technique allows the learned flow over $\mathcal{C}'$ to condition on superconducting temperature, allowing for property-conditioned inverse design. For CDVAE, one could fold T$_{\mathrm c}$ into the latent $\mathbf{z}$, add a head that recovers T$_{\mathrm c}$ from $\mathbf{z}$, and make the decoder explicitly conditional on the recovered T$_{\mathrm c}$ during denoising. This may be more robust than CDVAE’s current property-optimization approach (using an external predictor with latent-space gradient ascent) because conditioning on T$_{\mathrm c}$ is propagated throughout the architecture rather than confined to optimization of a random $\mathbf{z}$ using an external property predictor.

\end{document}